\documentclass[10pt,journal,compsoc,onecolumn]{IEEEtran}
\usepackage{cite}

\usepackage[cmex10]{amsmath}
\usepackage{amssymb}
\usepackage[linesnumbered,noline,boxed,commentsnumbered, ruled]{algorithm2e}
\usepackage{float}
\usepackage{graphicx}
\usepackage{subfig}
\usepackage{booktabs}
\usepackage{flushend}
\usepackage{multirow}
\usepackage{mathrsfs}
\usepackage{lscape}
\usepackage{bm}
\usepackage{tikz,pgf}
\usepackage{doi}
\usepackage{framed}
\usepackage{amsmath}

\usepackage[section]{placeins}
\usepackage{booktabs}
\usepackage{multirow}
\usepackage{longtable}
\usepackage{multicol}
\usepackage{smartdiagram}
\usepackage{genealogytree}
\usepackage{ragged2e}
\usepackage{colortbl}

\usetikzlibrary{arrows,shapes,positioning,shadows,trees}

\newcommand{\etal}{\textit{et al.}}

\definecolor{Ocean}{RGB}{129,194,234}

\begin{document}{\twocolumn}
	
	\title{What's the Situation with Intelligent Mesh Generation: A Survey and Perspectives}
	%
	%
	%
	%
	
	\author{Na~Lei,
		Zezeng~Li,
		Zebin~Xu,
		Ying~Li,
		and~Xianfeng Gu
		\IEEEcompsocitemizethanks{\IEEEcompsocthanksitem This research was supported by the National Key R$\&$D Program of China (2021YFA1003003) and the National Natural Science Foundation of China under Grants 61936002 and T2225012.
			
			\IEEEcompsocthanksitem N. Lei is with the International Information and Software Institute, Dalian University of Technology, Dalian, 116620, China (Email: nalei@dlut.edu.cn).
			
			\IEEEcompsocthanksitem Z. Li and Z. Xu are with the School of Software, Dalian University of Technology, Dalian, 116620, China (Email: zezeng.lee@gmail.com, xzb0516@mail.dlut.edu.cn).
			\IEEEcompsocthanksitem Y. Li is with the College of Computer Science and Technology, Jilin University, Changchun, 130015, China (Email: liying@jlu.edu.cn).
			\IEEEcompsocthanksitem X. Gu is with the Department of Computer Science and Applied Mathematics, State University of New York at Stony Brook, Stony Brook, NY 11794-2424, USA (Email: gu@cs.stonybrook.edu).
			
			\IEEEcompsocthanksitem Corresponding author: Ying~Li.}}
	
	%
	%


	\IEEEtitleabstractindextext{%
		\begin{abstract}
			\justifying
			Intelligent Mesh Generation (IMG) represents a novel and promising field of research, utilizing machine learning techniques to generate meshes. Despite its relative infancy, IMG has significantly broadened the adaptability and practicality of mesh generation techniques, delivering numerous breakthroughs and unveiling potential future pathways. However, a noticeable void exists in the contemporary literature concerning comprehensive surveys of IMG methods. This paper endeavors to fill this gap by providing a systematic and thorough survey of the current IMG landscape. With a focus on 113 preliminary IMG methods, we undertake a meticulous analysis from various angles, encompassing core algorithm techniques and their application scope, agent learning objectives, data types, targeted challenges, as well as advantages and limitations. We have curated and categorized the literature, proposing three unique taxonomies based on key techniques, output mesh unit elements, and relevant input data types. This paper also underscores several promising future research directions and challenges in IMG. To augment reader accessibility, a dedicated IMG project page is available at \url{https://github.com/xzb030/IMG_Survey}.
		\end{abstract}
		
		}

	\maketitle

	\IEEEdisplaynontitleabstractindextext
	\IEEEpeerreviewmaketitle

	\IEEEraisesectionheading{\section{Introduction}\label{sec:introduction}}
	\subsection{Motivation}
	\IEEEPARstart{I}{n} this paper, we embark on a systematic review of publications centered on Intelligent Mesh Generation (IMG). Recognized as one of the six fundamental research directions in NASA's Vision 2030~\cite{slotnick2014cfd}, mesh generation holds a pivotal role in computational geometry and is integral to numerical simulations. IMG amalgamates machine learning with mesh generation, a blend whose significance is gaining recognition as a growing corpus of research that explores the use of neural networks to generate high-quality meshes across diverse application scenarios. Driven by various meshing objectives, IMG research has seen a surge in interest. Notably, recent years have seen the advent of many innovative algorithms. Despite the substantial research on IMG, there is a conspicuous absence of comprehensive reviews that adopt a standardized methodology to ensure significance, completeness, and impartiality. To advance research in the IMG field, an urgent need exists for a systematic overview of the current state-of-the-art IMG methods. Such an overview will aid in distilling the essential commonalities in these works, identifying current research trends, and pinpointing promising future directions.
	
\begin{figure}[h!]
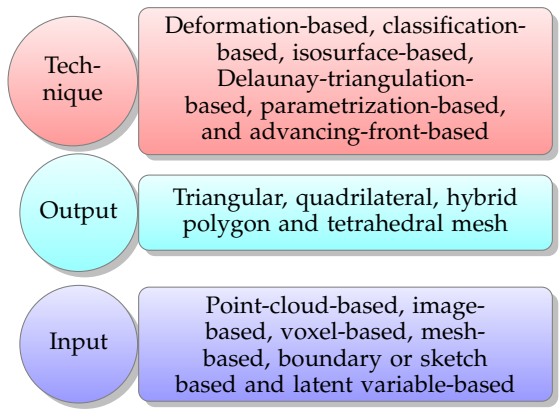

		\centering
		\smartdiagram[descriptive diagram]{
			{Technique, {Deformation-based, classification-based, isosurface-based, Delaunay-triangulation-based, parametrization-based, and advancing-front-based}},
			{Output, {Triangular, quadrilateral, hybrid polygon and tetrahedral mesh}},
			{Input, {Point-cloud-based, image-based, voxel-based, mesh-based, boundary or sketch based and latent variable-based}}}
		\caption{Taxonomies for existing IMG methods from three views.}
		\label{fig:allTaxonomy}
	\end{figure}

	Herein, our aim is to present a comprehensive and systematic survey, complete with a detailed taxonomy and content oriented evaluation. We undertake a multi-faceted analysis, examining key algorithmic techniques and their application scope, agent learning objectives, data types, targeted challenges, along with advantages and limitations. As depicted in Fig. \ref{fig:allTaxonomy}, we offer varied taxonomies for existing IMG methods, conceptualized from three perspectives: key techniques, output mesh unit elements, and relevant data types. Considering key techniques, we categorize IMG methods into deformation-based, classification-based, isosurface-based, Delaunay triangulation-based, parametrization based, and advancing front-based mesh generation. From the standpoint of output mesh unit elements, IMG methods are classified into triangular mesh, quadrilateral mesh, hybrid polygon mesh, and tetrahedral mesh generation. Lastly, we classify application input data types into point cloud-based, image-based, voxel-based, mesh-based, boundary or sketch-based, and latent variable-based mesh generation. Additionally, we present a concise overview of IMG's related areas, such as conventional neural networks for mesh learning, mesh quality metrics, and datasets. To summarize, we researched and reviewed 190 papers, which includes 113 papers focused on IMG and 77 papers on associated topics. Our paper makes four primary contributions:
	
	\begin{itemize}
		\item {
			We have undertaken a comprehensive review of 113 seminal IMG papers, evaluating them based on several key aspects: underlying techniques, the scope of application, agent learning objectives, data types, targeted challenges, and advantages and limitations.}
		\item 
		We have categorized existing IMG methods from three distinct vantage points: key techniques, output mesh unit elements, and suitable input data types;
		\item 
		We provide a broad overview of areas closely related to IMG, including conventional neural networks for mesh learning, mesh quality assessments, and datasets;
		\item 
		
		We encapsulate the present challenges confronting IMG and outline potential promising avenues that researchers could explore to address these challenges in their future endeavors.
	\end{itemize}
	\subsection{Related Surveys}
	A systematic literature review (SLR) \cite{keele2007guidelines}, as utilized in software engineering, is conducted by using a predefined methodical series of steps. To the best of our knowledge, no SLR has been conducted in the IMG domain. In this section, we describe some related review articles \cite{berger2017survey,han2019image,khan2020surface,xiao2020survey,nianhua2021preliminary}.
	
	Berger \etal \cite{berger2017survey} surveyed various point cloud-based surface reconstruction algorithms developed from 1992 to 2015. They classified the algorithms based on the type of priors that are employed, the ability to handle point cloud artifacts, input requirements, shape class, and reconstruction output. This review focuses on point cloud surface reconstruction whose output does not necessarily mesh. Note that mesh reconstruction and surface reconstruction are two different concepts. Mesh reconstruction refers to not only surface mesh but also volume mesh. However, surface reconstruction is not necessarily meshing but may also be voxel, RGBD image, signed distance field, etc. There have been other surveys covering only a specific surface reconstruction domain, such as image-based, 3D object reconstruction \cite{han2019image}, surface remeshing \cite{khan2020surface}, and advancing front-based methods \cite{nianhua2021preliminary}.
	All the surveys mentioned above focus on specific surface reconstruction domains but are not specialized for IMG. In addition, Xiao\cite{xiao2020survey} \etal. summarize geometric deep learning from a representational perspective. Thus, our paper is the first review of existing IMG methods.
	
	\section{Approach}\label{2}
	
	The aim of this paper is to offer a comprehensive and systematic survey by employing a dual-pronged strategy. In the initial phase, we delve into the existing literature, sieving out IMG-relevant work using both an open-ended interpretive approach and a keyword-focused literature review. The former approach is grounded on the authors' expertise, leading to a broad and exploratory collection of works, though it may introduce selection bias. In contrast, the latter targets specific IMG publications, offering a clear boundary, but its effectiveness relies heavily on the chosen keywords. In an effort to amalgamate the strengths of these two methods, we have subsequently applied both the keyword-centric and the open-ended interpretive approaches. In the subsequent phase, we compile and evaluate all articles gathered from the first phase based on the content extracted.
	
	\subsection{Search Strategy}
	The search strategy is paramount in assembling pertinent literature on a specific subject. The choice of the search string and the libraries are significant factors. We crafted a search string and selected five distinct libraries \footnote{
		1) IEEE Xplore \url{http://ieeexplore.ieee.org};
		2) ACM \url{https://dl.acm.org};
		3) ScienceDirect \url{http://www.sciencedirect.com};
		4) SpringerLink \url{https://link.springer.com};
		5) Wiley \url{https://onlinelibrary.wiley.com/}.} to scout for relevant papers. Adhering to the SLR guidelines \cite{keele2007guidelines}, we constructed a search string by amalgamating the keywords and their synonyms using Boolean operators.
	
	\noindent \textbf{Keywords and Boolean Operators:} [("computer graphics" OR "computational geometry") AND ("mesh reconstruction" OR "mesh generation" OR "Surface reconstruction" OR "3D object reconstruction" OR "triangulation" OR "quadrilateral" OR "tetrahedral" OR "hexahedral") AND ("intelligent" OR "learning" OR "data-based" OR "neural network" OR "ANN")] OR ("mesh evaluation" OR "3D object dataset" OR "mesh neural network" OR "mesh metric").
	
	\begin{figure*}[htbp!]
		\centering
		\vskip -0.2in
		\includegraphics[width=0.90\linewidth]{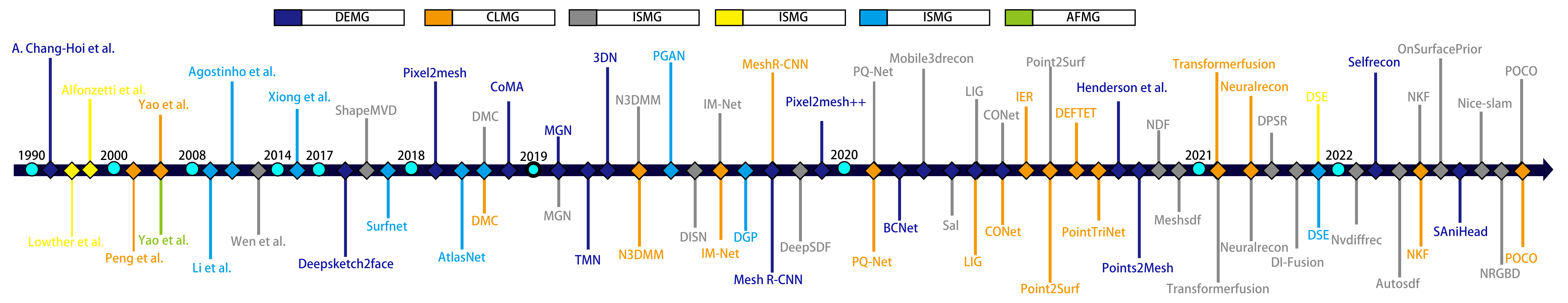}
		\vskip -0.1in
		\caption{Chronological overview of representative IMG methods}
		\label{fig:timeline}
	\end{figure*}
	\subsection{Content Extraction and Presentation of Findings}
	
	The existing literature offers a plethora of IMG techniques. We chose a total of 190 papers for a detailed analysis; 113 papers focus on IMG, while 77 papers delve into IMG's related areas, encompassing classical neural networks for mesh, mesh quality metrics, and mesh datasets, etc. In light of our research questions, we extracted the following details from each selected paper:
	1)The main challenges faced and significant breakthroughs achieved;
	2)The fundamental concept, scope of applicability, along with pros and cons;
	3)The type of input data, output mesh, and mesh quality.
	Furthermore, to appraise these IMG methods more objectively, we gathered the following additional information:
	1)The frequency of citation for the article (number of citations per month as per Google Scholar citation);
	2)The practicality of the algorithm proposed in the paper.

\begin{figure}[h]
		\centering
		\includegraphics[width=0.99\linewidth]{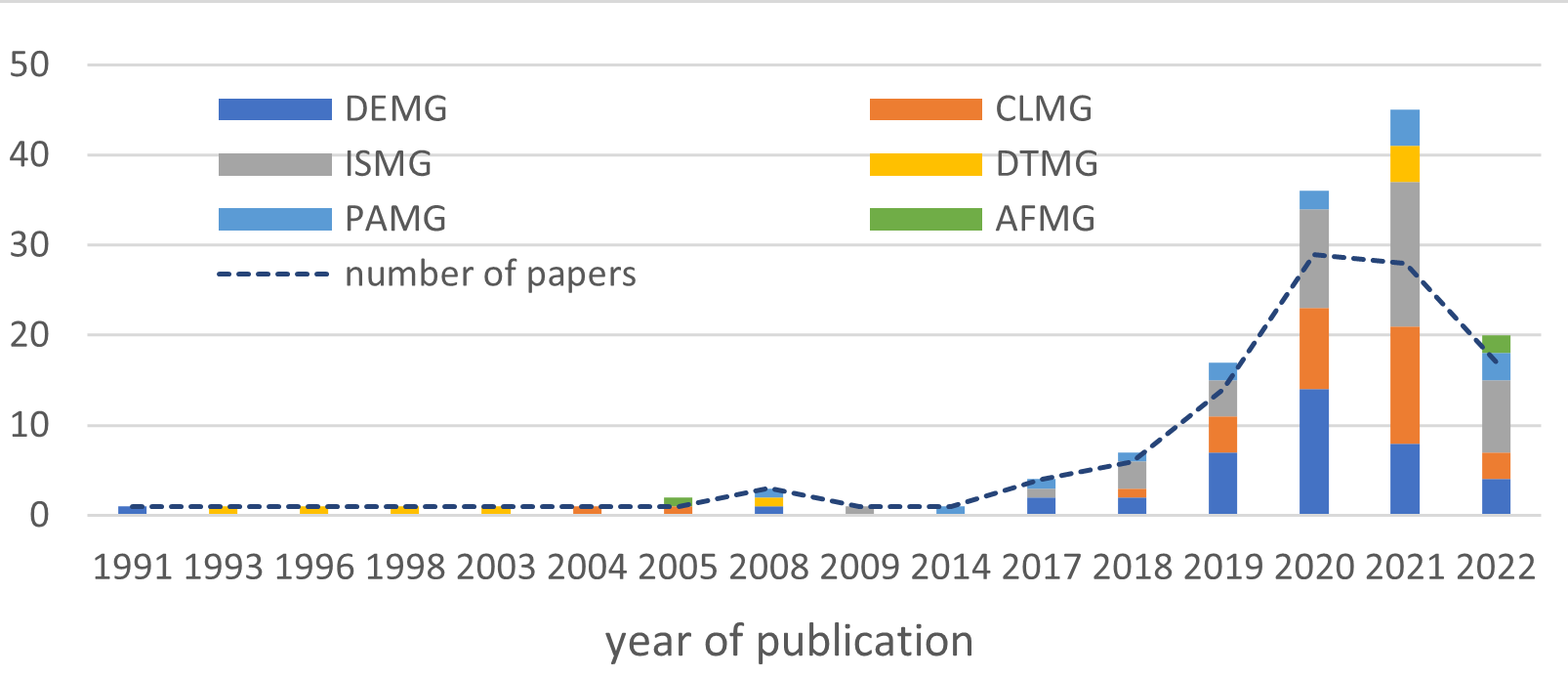}
		\caption{Annual distribution of articles of 6 technical types.}
		\label{fig:yeardistribution}
		\vskip -0.2in
	\end{figure}

	From the perspective of key techniques, output mesh types, and input data types, we first classified existing IMG papers and presented them in Table~\ref{tab:classfication}. This table also encapsulates whether the IMG is end-to-end, its principal network structure, and the learning goals of the neural network model. Table~\ref{tab:classfication} provides a clear categorization of the 113 IMG papers from three unique viewpoints, allowing an intuitive understanding of how intelligence is incorporated in each IMG method. Secondly, we delved into the detailed content of each IMG paper and organized Table~\ref{tab:procons} in chronological order. This table lists the targeted challenges, advantages, limitations, and average monthly citations for each study. We also showcased some representative methods in Fig. \ref{fig:timeline}, collated existing IMG papers into a line chart according to the number of papers published per year, and displayed this in Fig. \ref{fig:yeardistribution}. Given that our collection concluded in the first half of 2022, the number of IMG papers for the year has seen a notable decrease. However, Fig. \ref{fig:yeardistribution} illustrates that IMG research is currently at its peak. Furthermore, we displayed the number of papers published per year for each technology as a stacked bar chart. For papers belonging to multiple categories, we counted them separately; hence, the sum of bar values will exceed the total number of papers.
	
	\subsection{Fundamental Concepts}
	In general, meshes are divided into surface meshes and volume meshes. A surface mesh is a set of polygonal faces, and the goal is to form {  an approximation of an object's surface.} A polygonal surface mesh has three different combined elements: \emph{vertices}, \emph{edges}, and \emph{faces}. This mesh can also be considered the combination of \emph{geometry} and \emph{topology}, where the \emph{geometry} provides the positions of all its vertices, and the \emph{topology} provides the information between different adjacent vertices. Mathematically, a polygonal surface mesh $\mathcal{M}$ that contains $V$ vertices and $F$ faces is formed as
	\begin{equation}
		\label{eqn:01}
		\begin{aligned}
			\mathcal{M}=\{(\mathcal{V},\mathcal{F})&|\mathcal{V}=\{v_i\}_{i=1,2,...,V},\\
			&\mathcal{F}=\{f_i\}_{i=1,2,...,F},\ f_i\in\mathcal{V}^d\}\ ,\\
		\end{aligned}
	\end{equation}
	where $\mathcal{V}$ is the vertex set, $\mathcal{F}$ is the face set, and $d$ denotes the number of vertices of each facet. Similarly, a volume mesh $\mathcal{M}$ is formed as
	\begin{equation}
		\mathcal{M}=\left\{\left(\mathcal{V},\mathcal{F},\mathcal{T}\right)|\mathcal{T}=\{t_i\}_{i=1,2,...,T},\ t_i\in\mathcal{F}^k \right\},
	\end{equation}
	where $k$ represents the number of faces of each voxel $t_i$.
	
	To avoid ambiguity, the definition of mesh generation is given as follows: \\
	\textbf{Definition 1. Mesh Generation}:
	\textit{Mesh generation is such a task: given the input data $X$, the mapping $g: X\to\mathcal{M}$ acts on $X$ and outputs the surface mesh $\mathcal{M}=\{\mathcal{V},\mathcal{F}\}$ or volume mesh $\mathcal{M}=\{\mathcal{V},\mathcal{F},\mathcal{T}\}$.}
	
	In essence, the goal of mesh generation is to restore the missing geometry information or topology information in $X$. Through our survey, the input data types of existing IMG methods do not extend beyond point clouds, images, voxels, boundaries or sketches, meshes, and latent variables.
	
	\section{Classification based on Technique}\label{3}
	In this section, we categorize existing IMG methods into six groups, namely deformation-based, classification-based, isosurface-based, Delaunay triangulation-based, parametrization based, and advancing front-based mesh generation. This classification is grounded on the key techniques and guiding principles inherent in each method.
	
	In this classification, we illustrate how predominant IMG methodologies adeptly amalgamate traditional mesh generation techniques with novel deep learning modules, highlighting the merits and drawbacks of each approach. We anticipate this taxonomy will help readers grasp the function of neural networks in mesh generation more intuitively. In deformation-based IMG algorithms, neural network modules primarily predict vertex locations. For classification-based IMGs, neural networks primarily function as classifiers determining the connection or occupancy of vertices. Isosurface-based IMGs employ neural networks for implicit function fitting and zero isosurface extraction. In Delaunay triangulation-based IMGs, neural networks are variably employed to predict vertex coordinates or weights. For parameterization-based IMG algorithms, neural networks learn parameterized mapping or process parameterized data. In advancing front-based IMG algorithms, neural networks are utilized to predict the forward direction of the wave and the connection type of the new node. Further elaborations are provided in the corresponding subsections below.

	\subsection{Deformation-based Mesh Generation}\label{3.1}
\begin{figure}[htb]
		\centering
		\includegraphics[width=0.8\linewidth]{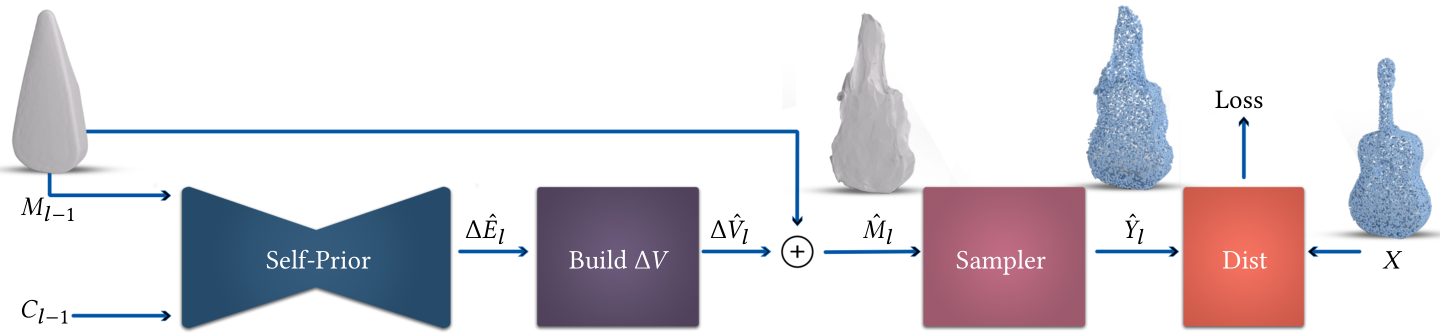}
		\caption{Point2Mesh \cite{hanocka_point2mesh_2020}: a deformation-based IMG method}
		\label{fig:deform}
		\vskip -0.1in
	\end{figure}

	A mesh consists of polygons and is essentially a discrete representation of a continuous surface. The combinatorial nature of the polygons prevents taking derivatives over the space of possible meshing of any given surface. Thus, it is difficult for mesh processing and optimization techniques to take advantage of modular gradient descent components of modern optimization frameworks. To circumvent this problem, deformation-based IMG has attracted more attention. As shown in Fig.\ref{fig:deform}, a distinguishing feature of this technology is the need for an initial mesh such as spherical or ellipsoidal template mesh. Deformation-based IMG is applicable to different data types of inputs. 
	A target mesh based on geometric information is usually generated from images \cite{danerek_deepgarment_2017,han2017deepsketch2face,wang2018pixel2mesh,bhatnagar2019multi,pan2019deep,venkat_humanmeshnet_2019,gkioxari_mesh_2019,wen2019pixel2mesh++,jiang2020bcnet,wang2020pixel2mesh,tong2020x,li20203d,dongsheng_3d_2020,zhang_meshingnet_2020,wang_surface_2020,henderson_leveraging_2020,yang2021lasr,wickramasinghe2021deep,smirnov2020learning,bertiche_deep_2021,li_learning_2021,pedone_learning_2021,hu_mesh_2021,wang_pixel2mesh_2021,tang2021skeletonnet,jiang2022selfrecon,ben_charrada_toponet_2022}, point clouds \cite{wang20193dn,rios_back_2020,hanocka_point2mesh_2020,gupta_neural_2020,ma_neural-pull_2021} or voxels\cite{wickramasinghe2020voxel2mesh,du_sanihead_2022}. 
	
	Due to the presence of an initial mesh, this type of method reduces the difficulty of mesh generation to some extent. The neural network only needs to predict the position of the vertex because the connection relationship already exists. The oldest deformation-based IMG work dates to 1991, and Ahn \etal\ \cite{chang-hoi_self-organizing_1991} proposed a self-organizing neural network to automatically generate a nonuniform density mesh. This is a pioneering work, albeit only plane meshes with simple 
	
	\begin{onecolumn}
		{
			\tiny
			\linespread{0.85} \selectfont
			\setlength{\tabcolsep}{3pt}
			\begin{longtable}{c|cccccc|cccccc|cccc|ccc}
				\caption{Taxonomy of the various types of IMG employed in the literature. DEMG, CLMG, ISMG, DTMG, PAMG and AFMG represent deformation-based, classification-based, isosurface-based, Delaunay-triangulation-based, parametrization-based, and advancing-front-based mesh generation, respectively. PC, IM, VO, ME, BS, and LA denote point-cloud-based, image-based, voxel-based, mesh-based, boundary or sketch-based, and latent variable-based IMG, respectively. Tri, Qua, Hyb and Tet represent triangular, quadrilateral, hybrid polygon and tetrahedral meshes, respectively. E2E means end-to-end, \checkmark indicates satisfaction.}\\
				\bottomrule 
				\multicolumn{1}{c}{\multirow{2}[4]{*}{\textbf{Article}}} &\multicolumn{6}{c}{\textbf{Technique}}
				&\multicolumn{6}{c}{\textbf{Input}} & \multicolumn{4}{c}{\textbf{Output}} & \multicolumn{3}{c}{\textbf{Attributes}}\\
				\cmidrule(r){2-7} \cmidrule(r){8-13} \cmidrule(r){14-17}  \cmidrule(r){18-20}
				\multicolumn{1}{c}{} & 
				\multicolumn{1}{c}{DEMG} & \multicolumn{1}{c}{CLMG} &  \multicolumn{1}{c}{ISMG} &\multicolumn{1}{c}{DTMG} & \multicolumn{1}{c}{PAMG} &\multicolumn{1}{c}{AFMG} & 
				\multicolumn{1}{c}{PC} & \multicolumn{1}{c}{IM} & \multicolumn{1}{c}{VO} &\multicolumn{1}{c}{ME}&  \multicolumn{1}{c}{BS} & \multicolumn{1}{c}{LA}   &\multicolumn{1}{c}{Tri} & \multicolumn{1}{c}{Qua} & \multicolumn{1}{c}{Hyb} & \multicolumn{1}{c}{Tet} &\multicolumn{1}{c}{E2E} &\multicolumn{1}{c}{Network Structure}  &\multicolumn{1}{c}{Learning Goals} \\
				\hline
				\endhead
				\rowcolor{Ocean} Self-organizing \cite{chang-hoi_self-organizing_1991} &\checkmark & & & & &     & & & & &\checkmark &       &\checkmark &\checkmark & &  &\checkmark &Self-organizing network  &Vertices location\\
				Lowther \etal \cite{lowther_density_1993} & & & &\checkmark & &     & & & & &\checkmark &       &\checkmark & & &   & &Feed forward net  &Element size\\
				\rowcolor{Ocean} Alfonzetti \etal \cite{alfonzetti_automatic_1996} & & & & \checkmark&&     & & & &\checkmark & &       &\checkmark & & &    & &Let-It-Grow ANN &Node positions\\
				Alfonzetti \etal \cite{alfonzetti_finite_1998} & & & & \checkmark& &     & & & &\checkmark & &       &\checkmark & & &     &&ANN&Node positions\\
				\rowcolor{Ocean} Alfonzetti \etal \cite{alfonzetti2003neural} & & & & \checkmark& &     & & & &\checkmark & &       &\checkmark & & &\checkmark &&ANN  &Node positions\\
				Peng \etal \cite{peng_3d_2004} & &\checkmark & & & &     & \checkmark& & & & &       &\checkmark & & &     &\checkmark&MLP &Coordinate $z$; point occupancy\\
				\rowcolor{Ocean} Yao \etal \cite{yao2005ann}& &\checkmark & & & &\checkmark    & & & & &\checkmark &     & &\checkmark & &    &\checkmark&ANN&Node positions\\
				Alfonzetti \etal \cite{alfonzetti_optimized_2008} & & & &\checkmark & &     & & & & \checkmark& &       &\checkmark & & &    & &Let-It-Grow ANN&Node positions\\
				\rowcolor{Ocean} Li \etal \cite{li_incomplete_2008} & & & & &\checkmark &     &\checkmark & & & & &       &\checkmark & & &     & &RBFs network&Radial basis function\\
				Agostinho \etal \cite{de_medeiros_brito_junior_adaptive_2008} & & & & &\checkmark &      &\checkmark & & & & &       &\checkmark & & &    &\checkmark&Self-organizing map &3D coordinates\\
				\rowcolor{Ocean} Wen \etal \cite{wen_rbf_2009} & & &\checkmark & & &     &\checkmark & & & & &       &\checkmark & & &     &&RBFs network&Radial basis function\\
				Xiong \etal \cite{xiong_robust_2014} & & & & & \checkmark&     &\checkmark & & & & &       &\checkmark & & &     &\checkmark&Sparse dictionary&Vertex position and triangulation\\
				\rowcolor{Ocean} DeepGarment \cite{danerek_deepgarment_2017} &\checkmark & & & & &     & &\checkmark & & & &       &\checkmark & & &       & \checkmark&SqueezeNet&Vertex positions\\
				Deepsketch2face \cite{han2017deepsketch2face} &\checkmark & & & & &    & & & & &\checkmark &      &\checkmark & & &      & \checkmark&AlexNet&Vertex positions\\
				\rowcolor{Ocean} ShapeMVD \cite{lun20173d} & & & \checkmark& & &     & & & & &\checkmark &      &\checkmark & & &    & &U-net&Depth and normal maps\\
				Surfnet \cite{sinha2017surfnet} & & & & &\checkmark &    & &\checkmark & & & &    &\checkmark & & &    & &Res-Unet& Geometry images\\
				\rowcolor{Ocean} Pixel2mesh \cite{wang2018pixel2mesh}&\checkmark & & & & &    & &\checkmark & & & &     &\checkmark & & &    &\checkmark&GCN;VGG-16;MLP & Vertex's coordinate;extraction characteristics\\
				AtlasNet \cite{groueix2018papier} & & & & &\checkmark &     &\checkmark &\checkmark & & & &       &\checkmark & & &   & & PointNet/ResNet+MLPs& Point position\\
				\rowcolor{Ocean} Li \etal  \cite{li2018robust} & & &\checkmark & & &     & & & & &\checkmark &       &\checkmark & & &    && U-Net& Normal, depth maps\\
				DMC \cite{liao_deep_2018} & &\checkmark &\checkmark & & &     &\checkmark & & & & &       &\checkmark & & &     &\checkmark&DMC &Occupancy and vertex displacement\\
				\rowcolor{Ocean} CoMA \cite{ranjan2018generating} &\checkmark & & & & &     & & & & & &\checkmark       &\checkmark & & &    & \checkmark& Autoencoder& Point position\\
				3D-CFCN \cite{cao2018learning} & & &\checkmark & & &     &\checkmark & & & & &       &\checkmark & & &   & & OctNet-based U-Net&Truncated signed distance field\\
				\rowcolor{Ocean} MGN  \cite{bhatnagar2019multi} &\checkmark & & & &\checkmark &    & &\checkmark & & & &    &\checkmark & & &   & \checkmark& MLP& Vertices position\\
				3DN \cite{wang20193dn} &\checkmark & & & & &     &\checkmark &\checkmark & & & &       &\checkmark & & &    & \checkmark & PointNet/VGG& Vertices position\\
				\rowcolor{Ocean} TMN  \cite{pan2019deep}&\checkmark & & & & &    & &\checkmark & & & &    &\checkmark & & &    & \checkmark& ResNet/MLP & Vertices position and errors\\
				ONet \cite{mescheder2019occupancy} & &\checkmark &\checkmark & & &     &\checkmark &\checkmark &\checkmark & & &       &\checkmark & & &     && ResNet/PointNet&Grid occupancy\\
				\rowcolor{Ocean} N3DMM \cite{bouritsas2019neural} &\checkmark & & & & &     & & & & & &\checkmark      &\checkmark & & &     &\checkmark&  Spiral-Conv GAN&Vertices position\\
				PGAN \cite{li_pgan_2019} & & & & &\checkmark &     & & & & & &\checkmark       &\checkmark & & &      && WGAN&Geometry image\\
				\rowcolor{Ocean} HumanMeshNet \cite{venkat_humanmeshnet_2019} & \checkmark& & & & &     & &\checkmark & & & &       &\checkmark & & &       & \checkmark&Resnet-18&Vertices position\\
				DISN \cite{xu_disn_2019} & & &\checkmark & & &     &\checkmark &\checkmark & & & &       &\checkmark & & &      && VGG-16 & Signed distance field\\
				\rowcolor{Ocean} IM-Net \cite{chen_learning_2019} & &\checkmark &\checkmark & & &     &\checkmark &\checkmark & & & &       &\checkmark & & &      && IM-Net &Signed distance field\\
				Scan2Mesh \cite{dai_scan2mesh_2019} & &\checkmark & & & &     & & &\checkmark & & &      &\checkmark & & &       & \checkmark& 3D-Conv GNN  &Mesh face\\
				\rowcolor{Ocean} DGP \cite{williams_deep_2019} & & & & &\checkmark &     &\checkmark & & & & &       &\checkmark & & &      & \checkmark &MLP &Local parametrization\\
				Mesh R-CNN \cite{gkioxari_mesh_2019} &\checkmark &\checkmark & & & &     & &\checkmark & & & &  &\checkmark & & &     & & Mesh R-CNN &Occupancy and point position\\
				\rowcolor{Ocean} DeepSDF \cite{park2019deepsdf} & & &\checkmark & & &     &\checkmark & & & & &       &\checkmark & & &      & & MLP  & Signed distance field\\
				Pixel2mesh++  \cite{wen2019pixel2mesh++} &\checkmark & & & & &   & &\checkmark & & & &      &\checkmark & & &    & \checkmark&VGG; GCN&Vertex position\\
				\rowcolor{Ocean} PQ-Net \cite{wu2020pq}& &\checkmark & \checkmark& & &   & &\checkmark & & & & \checkmark     &\checkmark & & &      & & Seq2Seq Autoencoder & Signed distance field\\
				BCNet \cite{jiang2020bcnet} &\checkmark & & & & &    & &\checkmark & & & &     &\checkmark & & &    & \checkmark&ResNet; GAT; Spiral-Conv&SMPL parameters; vertices position\\
				\rowcolor{Ocean}   PolyGen \cite{pmlr-v119-nash20a} &&\checkmark&&&& &&\checkmark&\checkmark&&& &&&\checkmark& &\checkmark&Transformer-based&Predict vertices and faces sequentially\\
				DGTS \cite{hertz2020deep} &\checkmark&&&&& &&&&\checkmark&& &\checkmark&&& &\checkmark &GCN &Displacement vector per face\\
				\rowcolor{Ocean}  Neural Subdivision \cite{liu2020neural}&\checkmark&&&&&    &&&&\checkmark&&  &\checkmark&&&  &\checkmark&MLP &Predict vertex position \\
				Mobile3drecon \cite{yang2020mobile3drecon} & & &\checkmark & & &   & &\checkmark & & & &    &\checkmark & & &   & &Res-UNet&Depth map \\
				\rowcolor{Ocean} Sal \cite{atzmon2020sal} & & &\checkmark & & &   &\checkmark & & & & &    &\checkmark & & &   &&MLP&Unsigned distance field\\
				Pixel2mesh2 \cite{wang2020pixel2mesh} &\checkmark & & & & &    & &\checkmark & & & &     &\checkmark & & &   & \checkmark&GCN;G-Resnet&Vertex position\\
				\rowcolor{Ocean} Voxel2mesh \cite{wickramasinghe2020voxel2mesh}&\checkmark & & & & &   & & &\checkmark & & &     &\checkmark & & &     & \checkmark&GCN-3D&Vertex position\\
				X-ray2shape \cite{tong2020x} &\checkmark & & & & &    & &\checkmark & & & &    &\checkmark & & &    &\checkmark&CNN;GCN&Vertex position\\
				\rowcolor{Ocean} Li \etal \cite{li20203d} &\checkmark & & & & &    & &\checkmark & & & &     &\checkmark & & &    &\checkmark&Resnet;3D-GCN&Vertex position\\
				Meshlet \cite{badki2020meshlet} & & &\checkmark & &\checkmark &    &\checkmark & & & & &     &\checkmark & & &    &&FC-based AE &Vertex positions\\
				\rowcolor{Ocean} LIG \cite{jiang2020local} & &\checkmark &\checkmark & & &     & \checkmark& & & & &       &\checkmark& & &    &&AE;3D-CNN;Residual blocks&Truncated signed distance field \\
				CONet \cite{peng2020convolutional} & &\checkmark &\checkmark & &      &   &\checkmark & & & & &      &\checkmark & & &    &&U-Net&Grid occupancy probability\\
				\rowcolor{Ocean} ILSM \cite{yan2020interactive} & &\checkmark &\checkmark & & &     & & & & &\checkmark &       &\checkmark & & &    & &CGAN&Occupancy; velocity field \\
				IER \cite{liu_meshing_2020} & &\checkmark & & & &     &\checkmark & & & & &       &\checkmark & & &    &\checkmark&SpareConv;MLP&Classifying triangles\\
				\rowcolor{Ocean} Yang \etal \cite{dongsheng_3d_2020} &\checkmark & & & & &     & &\checkmark & & & &       &\checkmark & & &   &\checkmark&VGG16;GCN;Graph attention&Vertex position\\
				MeshingNet \cite{zhang_meshingnet_2020} &\checkmark & & & & &     & & & & &\checkmark &       &\checkmark & & &    &\checkmark&FCN/Resnet&Vertex position\\
				\rowcolor{Ocean} Point2Surf \cite{erler_points2surf_2020} & & \checkmark&\checkmark & & &     &\checkmark & & & & &      &\checkmark & & &   &&MLP&Signed distance field\\
				DEFTET \cite{gao_learning_2020} & &\checkmark & & & &     &\checkmark &\checkmark & & & &       & & & &\checkmark    &&MLP;&Predict occupancy\\
				\rowcolor{Ocean} BTM \cite{rios_back_2020} &\checkmark & & & & &     &\checkmark & & & & &       &\checkmark & & &   &&Autoencoder&Initial mesh\\
				PointTriNet \cite{sharp_pointtrinet_2020} & &\checkmark & & & &     &\checkmark & & & & &       &\checkmark & & &    &\checkmark&MLP&Triangle label\\
				\rowcolor{Ocean} Surface Hof \cite{wang_surface_2020} &\checkmark & & & & &     & &\checkmark & & & &       &\checkmark & & &    &\checkmark&AE&Vertex position\\
				REIN \cite{daroya_rein_2020} & &\checkmark & & & &     &\checkmark & & & & &       &\checkmark & & &   &\checkmark&GraphRNN&Edge prediction\\
				\rowcolor{Ocean} Henderson \etal \cite{henderson_leveraging_2020} &\checkmark & & & & &     & &\checkmark & & & &   &\checkmark & & &   &\checkmark&CNN&Vertex position\\
				SSRNet \cite{mi_ssrnet_2020} & & &\checkmark & & &     &\checkmark & & & & &       & \checkmark& & &   &&UNet; tangent convolution&Octree vertex label\\
				\rowcolor{Ocean} MeshVAE \cite{yuan_mesh_2020} &\checkmark & & & & &     & & & & & &\checkmark   &\checkmark & & &   &\checkmark&CVAE; GCN&Vertex position\\
				Point2Mesh \cite{hanocka_point2mesh_2020} &\checkmark & & & & &     &\checkmark & & & & &     &\checkmark & & &    &\checkmark&GCN&Vertex position\\
				\rowcolor{Ocean} NMF \cite{gupta_neural_2020} &\checkmark & & & & &     &\checkmark &\checkmark & & & &       &\checkmark & & &    &\checkmark&Conditional Flow/PointNet&Vertex position\\
				NDF \cite{chibane2020neural}& & &\checkmark & & &    &\checkmark & & & & &   &\checkmark & & &    &&IF-Nets&Unsigned distance field\\
				\rowcolor{Ocean} Meshsdf \cite{remelli2020meshsdf}& & &\checkmark & & &   & &\checkmark & & & &    &\checkmark & & &   &\checkmark&SplineCNN&Vertices position\\
				Smirnov \etal \cite{smirnov2020learning} &\checkmark & & & &\checkmark &     & & & & &\checkmark &       & &\checkmark & &    &&ResNet-18&Local parameterization\\
				\rowcolor{Ocean} Lasr \cite{yang2021lasr}& \checkmark& & & & &    & & \checkmark & & & &   &\checkmark & & &    &\checkmark&Flow/Mask Nets&Vertices position\\
				Transformerfusion \cite{bozic2021transformerfusion}& &\checkmark &\checkmark & & &   & &\checkmark & & & &    &\checkmark & & &      &&Transformer; 3D-CNN&Learning occupancy field\\
				\rowcolor{Ocean} CSPNet \cite{venkatesh2021deep}& &\checkmark &\checkmark & & &    &\checkmark & & & & &   &\checkmark & & &    &&CONet; PointNet&Unsigned
				distance field\\
				DASM \cite{wickramasinghe2021deep}& \checkmark& & & & &   & &\checkmark& & & &    &\checkmark & & &    &\checkmark&GCN; Mesh R-CNN&Vertices position\\
				\rowcolor{Ocean} Neuralrecon \cite{sun2021neuralrecon}& &\checkmark &\checkmark& & &   & &\checkmark& & & &    &\checkmark & & &    &&GRU; MLP&Truncated signed distance field\\
				Sa-convonet \cite{tang2021sa}& &\checkmark &\checkmark & & &   &\checkmark & & & & &    &\checkmark & & &   &&PointNet; CNN&Learning occupancy fields\\
				\rowcolor{Ocean} DMTet \cite{NEURIPS2021_30a237d1}& & &\checkmark & & &   &\checkmark & &\checkmark & & &    &\checkmark & & &   &\checkmark&CGAN&Vertex position; signed distance field\\
				Vis2mesh \cite{song2021vis2mesh} & &\checkmark & &\checkmark & &    &\checkmark & & & & &     &\checkmark & & &  &&U-Net; PConv&visibility map\\
				\rowcolor{Ocean} IMLSNet  \cite{liu2021deep}& &\checkmark & \checkmark& &&     &\checkmark & & & & &       & \checkmark && &    & &Octree based CNNs&Signed distance field\\
				Retrievalfuse   \cite{siddiqui2021retrievalfuse}& & & \checkmark& &&     &\checkmark & & & & &       & \checkmark && &    &&U-net; Attention&Truncated distance field\\
				\rowcolor{Ocean} Deepdt   \cite{luo2021deepdt}& &\checkmark&\checkmark&&&     &\checkmark& & & & &       & \checkmark & & &    & &GCN&Inside/outside classification\\
				Iso-points   \cite{yifan2021iso}& & & \checkmark& &&     &\checkmark &\checkmark & & & &       & \checkmark & & &    &&IDR&Iso-surface\\
				\rowcolor{Ocean} DST \cite{rakotosaona2021differentiable} & & & &\checkmark & \checkmark     &    &\checkmark & & & & &       &\checkmark & & &  &\checkmark&--&Parameterization\\
				SAP \cite{peng2021shape}& & &\checkmark & & &     &\checkmark & & & & &      &\checkmark & & &   &&MLP  &Grid indicator function\\
				\rowcolor{Ocean} Bertiche \etal \cite{bertiche_deep_2021} & \checkmark& & & & &     & &\checkmark & & & &       &\checkmark & & &  &\checkmark&Point-CNN &SMPL parameters; vertices position\\
				NeeDrop \cite{boulch_needrop_2021} & & \checkmark&\checkmark & & &     &\checkmark & & & & &       &\checkmark & & &   &&ONet&Inside/outside classification\\
				\rowcolor{Ocean} LMR \cite{li_learning_2021} &\checkmark & & & & &     & &\checkmark & & & & &   \checkmark & & &    &\checkmark&RNN;&SMPL parameters \\
				NRSfM \cite{pedone_learning_2021} &\checkmark & & & & &     & &\checkmark & & & &       & &\checkmark & &   &\checkmark&Unet&Location of points\\
				\rowcolor{Ocean} AnalyticMesh \cite{lei_learning_2021} & & \checkmark&\checkmark & & &     &\checkmark & & & & &       & & &\checkmark &    &\checkmark&MLPs&Isosurface\\
				DHSP \cite{wei_deep_2021} & \checkmark& & & &\checkmark &     &\checkmark & & & & &       &\checkmark & & &    &&MeshCNN;2DCNN&Vertices position; texture map\\
				\rowcolor{Ocean} Neural-Pull \cite{ma_neural-pull_2021} & & &\checkmark & & &     &\checkmark & & & & &       &\checkmark & & &   & &MLP&Signed distance field\\
				DI-Fusion \cite{huang_di-fusion_2021} & & &\checkmark & & &     & &\checkmark & & & &       &\checkmark & & &   &&MLP based AE&Truncated signed
				distance field\\
				\rowcolor{Ocean} DSE \cite{rakotosaona_learning_2021} & & & &\checkmark & \checkmark&     &\checkmark & & & & &   &\checkmark & & &    &&FoldingNet&Parameterization \\
				LDFQ \cite{dielen_learning_2021} & & & & &\checkmark &     & & & &\checkmark & &       & &\checkmark & &  &&Pointnet;SpiralNet;&Direction fields \\
				\rowcolor{Ocean} DGNN \cite{sulzer_scalable_2021} & &\checkmark & &\checkmark & &     &\checkmark & & & & &       & \checkmark& & &   &\checkmark&GNN+MLP&Tetrahedron inside/outside scores\\
				Hu \etal \cite{hu_mesh_2021} &\checkmark & & & & &     & &\checkmark & & & &       &\checkmark & & &   &\checkmark&GAN&Mesh texture\\
				\rowcolor{Ocean} NMC \cite{chen_neural_2021} & &\checkmark &\checkmark & & &     & & &\checkmark & & &       &\checkmark & & &    &\checkmark&MLP&Isosurface\\
				Skeletonnet \cite{tang2021skeletonnet}& \checkmark& \checkmark&\checkmark & & &   & &\checkmark & & & &     &\checkmark & & &  &&Point2voxel; 3D-Unet&Inside/outside classification\\
				\rowcolor{Ocean} Nvdiffrec \cite{munkberg2022extracting} & & & \checkmark& & &   & &\checkmark & & & &    &\checkmark& & &      &\checkmark &GAN;MLP &Signed distance field; texture \\
				Selfrecon \cite{jiang2022selfrecon}&\checkmark& & & & &   & &\checkmark & & & &    &\checkmark & & &    &&MLP&Point location and isosurface\\
				\rowcolor{Ocean} Autosdf \cite{mittal2022autosdf}& & &\checkmark & & &   & &\checkmark & & & &    &\checkmark& & &   & &VQ-VAE; transformer&Signed distance field \\
				NKF \cite{williams2022neural}& & \checkmark&\checkmark & & &   & \checkmark& & & & &    &\checkmark & & &    &&CNN &Inside/outside classification\\
				\rowcolor{Ocean} Sketch2PQ \cite{deng_sketch2pq_2022} & & & & &\checkmark &     & & & & &\checkmark &       & &\checkmark & &   &\checkmark& U-Net&Spline surfaces and directional fields\\
				SRMAE \cite{hahner_mesh_2022} &\checkmark & & & & &     & & & &\checkmark & &       &\checkmark & & &   &\checkmark&AE; 2DConv& Hexagonal grids position\\
				\rowcolor{Ocean} OnSurfacePrior \cite{ma2022reconstructing} & & &\checkmark & & &     &\checkmark  & & & & &       &\checkmark & & &   &&MLP&Signed distance field\\
				Lu \etal \cite{lu_new_2022} & & & & & &\checkmark     & & & & &\checkmark &       & & &\checkmark &    &\checkmark&BP-ANN&Node position; angle \\
				\rowcolor{Ocean} MGNet \cite{chen_mgnet_2022} & & & & &\checkmark &     & & & & &\checkmark &      & &\checkmark & &   &\checkmark&MLPs&Points location\\
				RLQMG \cite{pan_reinforcement_2022} & & & & & &\checkmark     & & & & &\checkmark &      & &\checkmark & &    &\checkmark&RL; FNN&Points location; insert type\\
				\rowcolor{Ocean} TopoNet \cite{ben_charrada_toponet_2022} &\checkmark & & & & &     & &\checkmark & & & &      &\checkmark & & & &\checkmark&pixel2mesh; GCN; RL&Points location; facet occupancy\\
				SAniHead \cite{du_sanihead_2022} &\checkmark & & & & &     & & & & &\checkmark &      &\checkmark & & &  &\checkmark&Pixel2mesh; GCN&Points location\\
				\rowcolor{Ocean} NDC \cite{chen_neural_2022} & & \checkmark&\checkmark & & &     &\checkmark & &\checkmark& & &     &\checkmark & & &   &\checkmark&3D CNN/pointnet++ &Predict edges\\
				PCGAN\cite{li_pcgan_2022} & & & & &\checkmark &     & & & & & &\checkmark      &\checkmark & & &   &&GAN&Geometric images \\
				\rowcolor{Ocean} Nice-slam \cite{zhu2022nice} & & &\checkmark & & &     & &\checkmark & & & &      &\checkmark & & &    &&MLPs&Isosurface\\
				NRGBD\cite{azinovic2022neural}& & &\checkmark & & &   & &\checkmark & & & &     &\checkmark & & &   &&NeRF&Signed distance field\\
				\rowcolor{Ocean} POCO \cite{boulch2022poco}& & \checkmark&\checkmark & & &   & \checkmark& & & & &     &\checkmark & & &   & & FKAconv; Attention model& Point Occupancy 
				\label{tab:classfication}\\
				
			\end{longtable}
		}
	\end{onecolumn}
	\twocolumn
	
	\noindent geometry. In recent years, a sheer volume of practical methods has emerged. Most of the early deformation-based IMG works involved the deformation of the basic model to obtain the target mesh, such as clothing \cite{danerek_deepgarment_2017}, face \cite{han2017deepsketch2face}, and body \cite{venkat_humanmeshnet_2019}. These methods only generate deformation meshes of specific target objects. Later, a series of algorithms represented by Pixel2Mesh \cite{wang2018pixel2mesh} introduced templates as the basic mesh, which greatly improved the practicality and generalization. The deformation-based approach is tightly integrated with GCN and MLP for better prediction of vertex positions.
	
	However, all deformation-based models have a common drawback: they only generate meshes with the same topology as the initial mesh. Therefore, another research direction is how to adapt it to a variety of topologies. Rios \etal\ \cite{rios_back_2020} proposed a very direct solution by providing multiple basic template meshes for the algorithm to automatically select. Notably, this method does not fundamentally solve the above problems but only alleviates them. To proceed, Charrada \etal\ \cite{ben_charrada_toponet_2022} introduced a face-pruning mechanism, which iteratively adapted the topology of the mesh, using face-pruning operations, while keeping the main properties of the template, i.e., the appealing visual aspect and uniform mesh connectivity. Admittedly, the face-pruning mechanism can make the method adaptable to mesh generation for more complex topologies.
	\subsection{Classification-based Mesh Generation}
\begin{figure}[htb]
		\centering
		\includegraphics[width=0.76\linewidth]{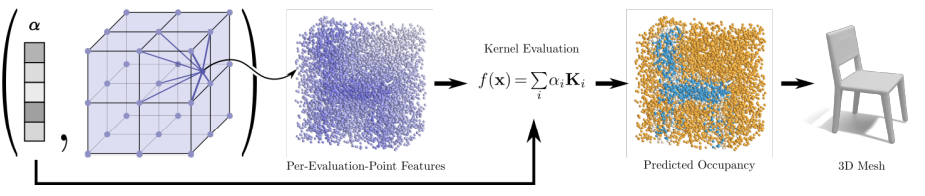}
		\caption{NKF \cite{williams2022neural}: an occupancy classification-based IMG.}
		\label{fig:occupy}
		\vskip -0.15in
	\end{figure}

\begin{figure}[htb]
		\centering
		\includegraphics[width=0.76\linewidth]{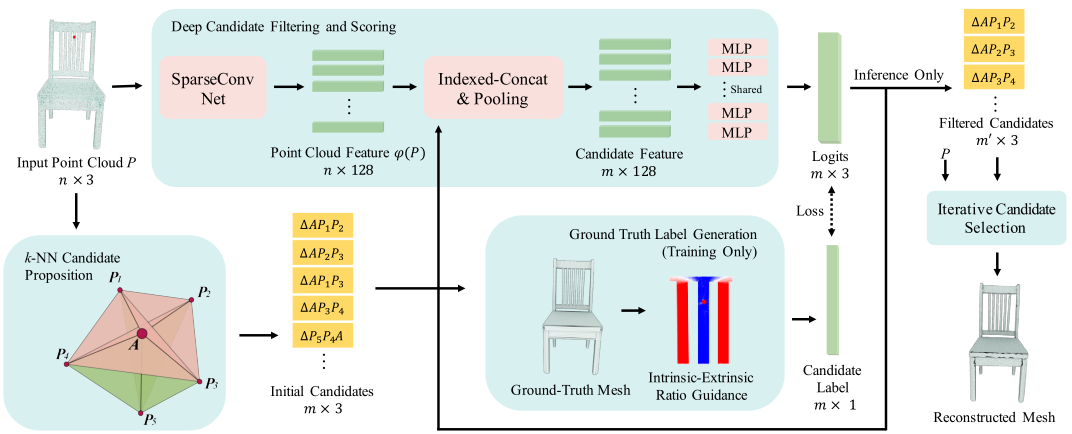}
		\caption{IER \cite{liu_meshing_2020}: a facet existence classification-based IMG.}
		\label{fig:classify}
	\end{figure}

	Inspired and encouraged by the classification model in the structured data domain, many researchers have designed various effective IMG algorithms by combining classification model-based neural networks with mesh generation. These algorithms are divided into two categories: occupancy prediction \cite{peng_3d_2004,liao_deep_2018,mescheder2019occupancy,chen_learning_2019,gkioxari_mesh_2019,wu2020pq,jiang2020local,peng2020convolutional,yan2020interactive,erler_points2surf_2020,gao_learning_2020,bozic2021transformerfusion,venkatesh2021deep,sun2021neuralrecon,tang2021sa,liu2021deep,luo2021deepdt,boulch_needrop_2021,sulzer_scalable_2021,chen_neural_2021,lei_learning_2021,song2021vis2mesh,williams2022neural,chen_neural_2022,boulch2022poco,tang2021skeletonnet} and mesh basic element prediction \cite{dai_scan2mesh_2019,liu_meshing_2020,daroya_rein_2020,sharp_pointtrinet_2020}. The former voxelized the three-dimensional space and decides whether the voxel belongs to an object, which is a binary classification problem. Alternatively, occupancy prediction can be further regarded as the three classification problems of the interior, surface and exterior of the object. The latter regards mesh generation as a combination of basic elements, so the generation model only needs to judge whether the basic elements to be detected should exist, which is also a binary classification problem in essence. Examples include the classification of facet existence and local connectivity. Besides binary classification, multiclassification can also be designed to generate meshes, such as IER \cite{liu_meshing_2020}. Fig. \ref{fig:occupy} and Fig. \ref{fig:classify} show the IGM methods based on occupancy and basic element classification, respectively.
	
	IMG algorithms based on occupancy prediction are represented by occupancy networks \cite{mescheder2019occupancy}, which are mainly used to address the task of reconstructing mesh by giving single or multiple images. Early works locate the object surface by judging the occupancy of each grid and then reconstruct the mesh by marching cubes \cite{lorensen1987marching}. In the high-resolution case, the floating point operation proportional to the cube of the voxel resolution is unbearable. The grid occupied by the object often consists of only a small number of cubes gathered in the same area, which means that many calculations are wasted. To overcome the problem of high computational complexity, IM-NET \cite{chen_learning_2019} proposed a more efficient sampling approach that samples more points near shape surfaces and disregards most points far away. When addressing the task of generating meshes from point clouds, subsequent works \cite{erler_points2surf_2020,boulch2022poco} focused on judging the occupancy of the point cloud instead of the whole 3D space, which directly reduces the floating-point computation. 
	
	Compared with the former, these IMG methods based on mesh basic element prediction generate meshes end-to-end. Yao \cite{yao2005ann} proposed the first algorithm to generate a mesh using a classification model, which is employed to determine whether to add new vertices and in what way. IER \cite{liu_meshing_2020} proposed a set of
	candidate triangle faces by constructing a k-nearest neighbor graph on the input point cloud, where the neural network was utilized to filter out the incorrect candidates. The flowchart of IER is provided in Fig. \ref{fig:classify}. Scan2Mesh \cite{dai_scan2mesh_2019} and REIN \cite{daroya_rein_2020} generated meshes by classifying the existence of edges. PointTriNet \cite{sharp_pointtrinet_2020} iteratively applied two neural networks: a classification network predicts whether a candidate triangle should appear in the mesh, while a proposal network suggested additional candidates. The proposal network of PointTriNet avoids the need to define candidate faces in advance and removes the assumption that all points must be derived from triangular mesh vertices. Although the application of PointTriNet has expanded, the generated mesh still has many problems, such as nonmanifold and many holes.
	\subsection{Isosurface-based Mesh Generation}
\begin{figure}[htb]
		\centering
		\includegraphics[width=0.4\linewidth]{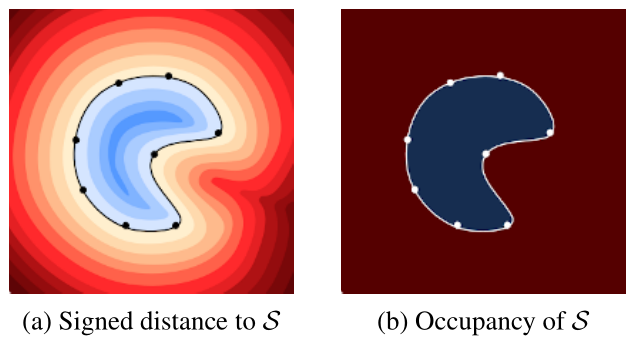}
		\includegraphics[width=0.4\linewidth]{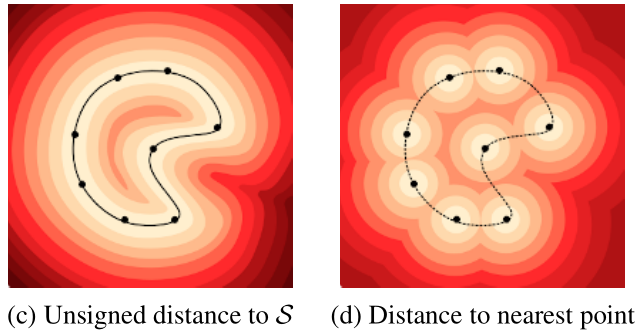}
		\caption{Different types of implicit functions \cite{boulch_needrop_2021}.}
		\label{fig:implicits}
	\end{figure}

	It is assumed that each point $X\in{  \mathbb{R}^3}$ has an attribute value and that $H: X\to {  \mathbb{R}}$ denotes implicit functions in space $X$. Then, the surface composed of continuous spaces with equal attribute values is referred to as an isosurface, which is defined as:
	\begin{equation}
		\label{eqn:Isosurface}
		Isosurface(h)=\{X|H(X)=h\}\\
	\end{equation}
	Existing works on isosurface-based IMG leverage neural networks to fit implicit functions and extract $\emph{Isosurface(0)}$. Triangle meshes are usually reconstructed by marching cubes \cite{lorensen1987marching}, marching tetrahedra \cite{doi1991efficient}, Poisson surface reconstruction \cite{kazhdan2013screened} algorithms or neural networks \cite{liao_deep_2018,lei_learning_2021,NEURIPS2021_30a237d1,chen_neural_2021,chen_neural_2022}. According to the different selected implicit functions, these IMG methods are divided into four categories: radial basis functions (RBFs) \cite{wen_rbf_2009}, occupancy fields \cite{lun20173d,liao_deep_2018,li2018robust,mescheder2019occupancy,yan2020interactive,mi_ssrnet_2020,yang2020mobile3drecon,tang2021skeletonnet,bozic2021transformerfusion,zhu2022nice,tang2021sa,williams2022neural,boulch2022poco}, signed distance functions (SDFs) \cite{cao2018learning,xu_disn_2019,chen_learning_2019,erler_points2surf_2020,wu2020pq,jiang2020local,peng2020convolutional,remelli2020meshsdf,siddiqui2021retrievalfuse,luo2021deepdt,NEURIPS2021_30a237d1,liu2021deep,yifan2021iso,lei_learning_2021,ma_neural-pull_2021, sun2021neuralrecon,huang_di-fusion_2021, peng2021shape,chen_neural_2021,ma2022reconstructing,chen_neural_2022,munkberg2022extracting,mittal2022autosdf,azinovic2022neural}, and unsigned distance function (UDFs) \cite{atzmon2020sal,chibane2020neural,venkatesh2021deep,boulch_needrop_2021}. The different types of implicit functions are shown in Fig. \ref{fig:implicits}.
	
	Isosurface-based IMG has the benefit of representing complex geometry and topology and is not limited to a predefined resolution. Therefore, isosurface-based IMG is favored by researchers. In 2009, Wen \etal \cite{wen_rbf_2009} utilized the least squares radial basis function-based neural network to estimate the coefficient on each surface sample and constructed a continuous implicit function to represent a 3D surface. The network also overcame the problem of numerical ill-conditioning and overfitting of traditional RBF reconstruction and offered a tool to utilize fewer samples to reconstruct models and make geometry processing feasible and practical. Afterward, {  Lun \etal \cite{lun20173d} used a decoder network to fit the foreground probability function with the occupancy fields}; it is defined as follows:
	\begin{equation}
		\centering
		\label{eq:occupancy}
		O: X\to \{0,1\},\ X{  \in\mathbb{R}^3}\ .
	\end{equation}
	
	Recently, many works based on occupancy prediction emerged. Liao \etal \cite{liao_deep_2018} proposed differentiable Marching Cubes by predicting the probability of occupancy for each
	voxel. To improve voxel occupancy prediction's calculation and memory efficiency, Mescheder \etal proposed occupancy networks \cite{mescheder2019occupancy}. Currently, various occupancy-based methods with different network architectures and training strategies have been proposed to continuously improve the efficiency, robustness, and accuracy \cite{li2018robust,yan2020interactive,mi_ssrnet_2020,yang2020mobile3drecon,tang2021skeletonnet,bozic2021transformerfusion,zhu2022nice,tang2021sa,williams2022neural,boulch2022poco}. The other two commonly utilized implicit functions are 
	SDF and UDF. Cao \etal \cite{cao2018learning} introduced a cascaded, 3D, convolutional network architecture that learned SDF from noisy and incomplete depth maps in a progressive, coarse-to-fine manner. Wang \etal \cite{xu_disn_2019} presented a deep implicit surface network that generated a high-quality, detail-rich, 3D mesh from a 2D image by predicting the underlying SDF. Among these methods, IMG methods of utilizing  Neural Radiance Fields (NeRF) show promising results in reproducing the appearance of an object or scene \cite{wangneus,azinovic2022neural,wang2022go-surf}. For them, NeRF is used to predicate SDF or density fields. Then the mesh can be extracted from the SDF using the marching cubes algorithm.
	
	Although neural implicit representations gained popularity in 3D reconstruction due to their expressiveness and flexibility, the nature of neural implicit representations produces slow inference time and requires careful initialization. To solve these problems, Peng \etal \cite{peng2021shape} introduced a differentiable point-to-mesh layer using a differentiable PSR \cite{kazhdan2013screened}, which bridged the explicit 3D point representation with the 3D mesh via the implicit indicator field, enabling end-to-end optimization.
	\begin{figure}[htb!]
		\centering
		\includegraphics[width=0.75\linewidth]{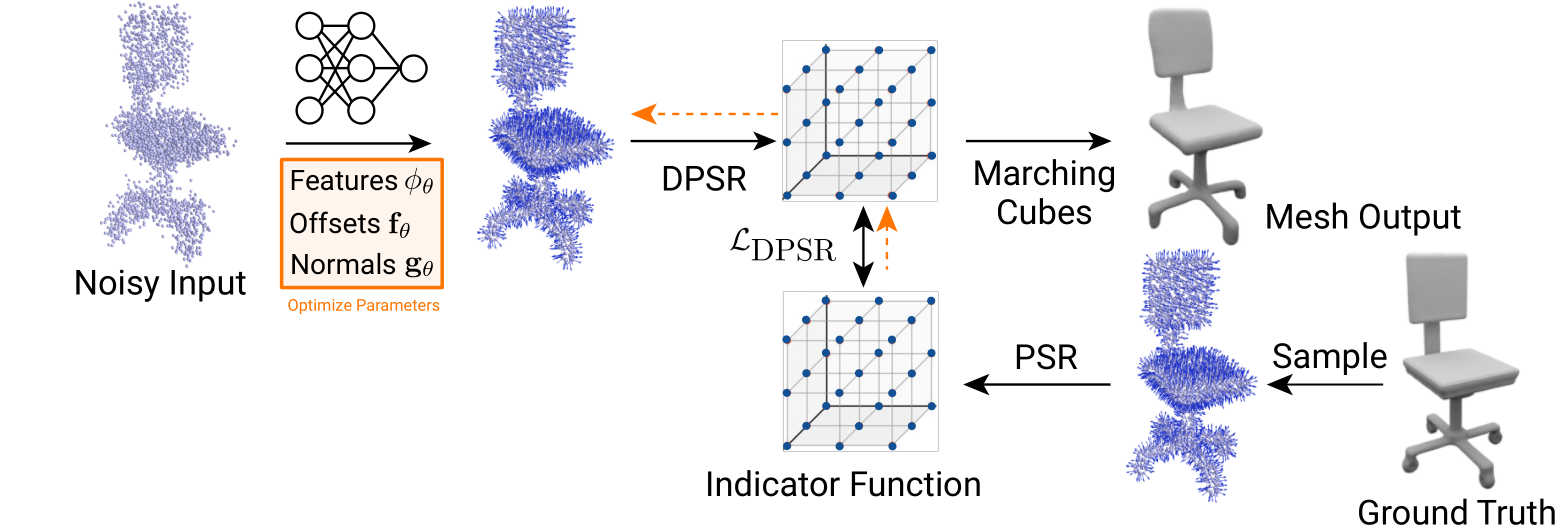}
		\caption{SAP \cite{peng2021shape}: an isosurface-based IMG method}
		\label{fig:Isosurface}
	\end{figure}
	Its flowchart is shown in Fig. \ref{fig:Isosurface}. For UDF, Atzmon \etal \cite{atzmon2020sal} introduced a sign-agnostic learning approach for learning implicit shape representations directly from raw and unsigned geometric data. In \cite{chibane2020neural}, neural distance fields were proposed to predict the unsigned distance field for arbitrary 3D shapes given sparse point clouds. To estimate local geometric such as the normals and tangent planes by implicit representation,  Venkatesh \etal \cite{venkatesh2021deep} utilized an efficient implicit representation referred to as the closest surface point representation, which was utilized to solve the PDE-related problems on surfaces\cite{ruuth2008simple,macdonald2008level,macdonald2010implicit,Calculus,5414447,MACDONALD20117944}. To process highly sparse input point clouds, NeeDrop \cite{boulch_needrop_2021} introduced a statistically based self-supervised approach to estimate occupancy functions directly from point clouds.
	
	The family of isosurface approaches is sometimes
	limited by their sensitivity to noise, outliers, and nonuniform sampling. In addition, solving equations on a large scale in implicit methods can be time-consuming. There is a cubic or quadratic relation between resolution and run time as well as memory usage, which limits its application value.
	\subsection{Delaunay triangulation-based Mesh Generation}
	Delaunay triangulation \cite{lee1980two} is a widely employed mesh reconstruction technology that can connect point clouds into triangular meshes. Delaunay triangulation has many excellent properties, such as maximizing the minimum angle characteristic. Regardless of which region the Delaunay triangulation is constructed from, the final generated triangular mesh is unique. Inspired by Delaunay triangulation, some researchers have tried to integrate it into IMG to generate high-quality and manifold meshes \cite{lowther_density_1993,alfonzetti_automatic_1996,alfonzetti_finite_1998,alfonzetti2003neural,alfonzetti_optimized_2008,rakotosaona2021differentiable,sulzer_scalable_2021,rakotosaona_learning_2021}.
	\begin{figure}[htb!]
		\centering
		\includegraphics[width=0.8\linewidth]{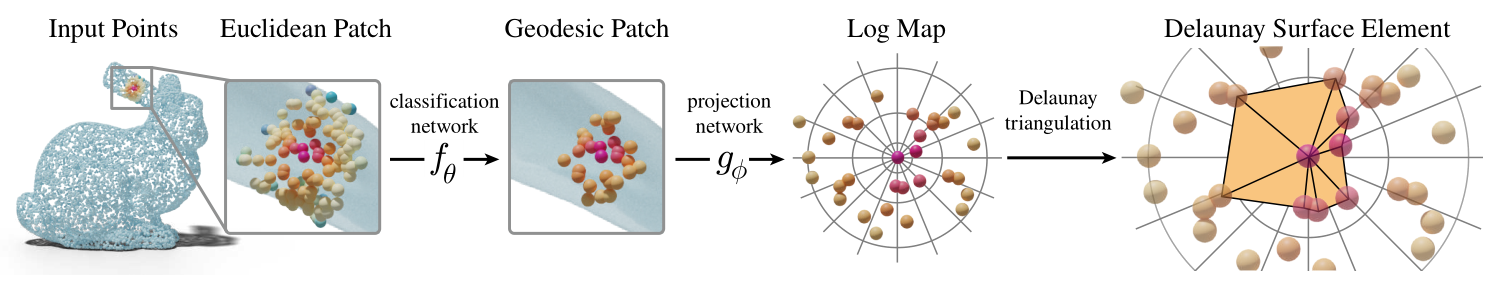}
		\caption{DSE \cite{rakotosaona_learning_2021}: a Delaunay triangulation-based IMG}
		\label{fig:Delaunay}
	\end{figure}
	Among these methods, early works \cite{lowther_density_1993,alfonzetti_automatic_1996,alfonzetti_finite_1998,alfonzetti_optimized_2008} predict the position of vertices using neural networks and then generate high-quality meshes using Delaunay triangulation \cite{lee1980two}. These methods generate 2D planar meshes with given boundaries or coarse meshes. However, in applications, a three-dimensional surface often needs to be addressed. Alfonzetti \cite{alfonzetti2003neural} utilized the 3D Delaunay algorithm to generate a tetrahedral mesh from a 3D point cloud. Song \etal\ \cite{song2021vis2mesh} combined the classification model and adaptive Delaunay algorithm to realize the mesh reconstruction of large indoor and outdoor scenes. Rakotosaona \etal\ \cite{rakotosaona_learning_2021} projected a 3D point into the 2D parameter space by a neural network, triangulated the 2D plane point using the Delaunay algorithm, and then pulled the point cloud back to the 3D surface by inverse mapping. The algorithm pipeline is shown in Fig. \ref{fig:Delaunay}. Furthermore, to realize end-to-end differentiable triangulation, Rakotosaona \etal\ \cite{rakotosaona2021differentiable} proposed a differentiable weighted Delaunay triangulation.
	
	\subsection{Parametrization-based Mesh Generation}\label{3.5}
	\begin{figure}[htb]
		\centering
		\includegraphics[width=0.8\linewidth]{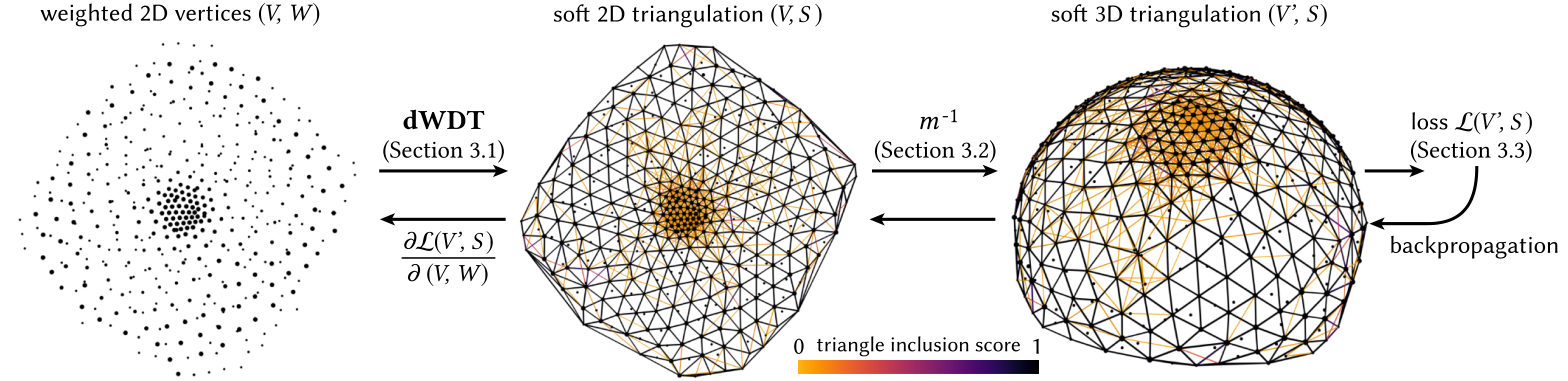}
		\caption{DST \cite{rakotosaona2021differentiable}: a parametrization-based IMG method}
		\label{fig:Parametric}
		\vskip -0.1in
	\end{figure}
	As \cite{ray2006periodic} remark, parameterization is a correspondence between a surface mesh embedded in 3D and a simple 2D domain referred to as the parameter space. Generally, parameterization is expected to be bijective, which is a classic problem in computer graphics and geometry processing with multifarious applications, such as mesh generation, texture mapping, and shape correspondence. Considering the convenience of mesh generation in parameter space, some researchers have tried to combine parametrization with machine learning modules and then invented a series of IMG methods based on parameterization. According to the different participation modes of parameterization, we broadly divide them into two categories. The \textbf{first category} is geometry image-based IMG. The authors map the surface to the parameter plane by the traditional parametric method \cite{gu2002geometry} to obtain a geometry image, process the geometry image using the deep learning model, and then convert the geometry image to a mesh. The \textbf{second category} is parameterization learning-based IMG, which learns the mapping from the parameter space to the target space. Then, meshes are generated in the parameter space by the traditional or machine learning method.
	
	For the \textbf{first category}, Sinha
	\etal\ \cite{sinha2017surfnet} developed a procedure to create geometry images that represent the shape surface of a category of 3D objects. Then, the authors utilized these geometry images for category-specific surface generation by developing a variant of deep residual networks. Li \etal\ \cite{li_pgan_2019,li_pcgan_2022} proposed a prediction generative adversarial network and prediction-compensation generative adversarial network to learn a joint distribution of geometry and normal images for generating meshes. They aimed to generate meshes with two generative adversarial networks (GANs) \cite{goodfellow2020generative} and achieved good performance in face mesh generation. The advantage of this type of approach is that 3D objects can be directly processed by 2D grid convolution. However, angle and area distortion is often disregarded. Surfnet \cite{sinha2017surfnet} reduced the effect of area distortion to some extent by making different objects of the same class have the same geometry image. For the \textbf{second category}, the main idea was to fit a mapping from the parameter space to the object space or vice versa by a neural network. Early work \cite{li_incomplete_2008} promoted the planar mesh to the surface by fitting the bivariate function. In \cite{de_medeiros_brito_junior_adaptive_2008}, the mapping from a 2D triangular lattice to a surface in 3D space is learned to lift the mesh in parameter space to the target surface. Other researchers try to project the 3D point cloud onto a 2D plane, generate meshes on the plane, and then pull it back. Considering the complexity and difference in surface topology, part of the work chooses to fit local homeomorphic mapping \cite{groueix2018papier,williams_deep_2019,badki2020meshlet,smirnov2020learning,rakotosaona_learning_2021,dielen_learning_2021,rakotosaona2021differentiable}. The other part directly fits the global mapping by assuming the surface type \cite{xiong_robust_2014,wei_deep_2021,deng_sketch2pq_2022,chen_mgnet_2022}. Fig. \ref{fig:Parametric} shows a local parameter learning method, DST \cite{rakotosaona2021differentiable}, which decomposed the source into local patches and then performed per-patch meshing. The method employed parameterization $m$, a bijective and piecewise differentiable mapping, to project 3D surfaces into 2D parametric space. Then, the 2D vertices were lifted to form a soft 3D triangulation on the surface using $m^{-1}$. The advantage of DST is to ensure that the generated mesh is 2-manifold, and the limitation is the boundary artifact caused by partitioning, which is a common problem for all local parameterization methods.
	
	\subsection{Advancing front-based Mesh Generation}
	The advancing front method is a greedy algorithm that gradually generates mesh nodes from the boundaries to the interior and recursively executes. Each recursive process is divided into three steps: first, a line segment is selected from the generation segment front set that splits the meshed domain and unmeshed domain, where the selected segment is referred to as the base segment, as it will form the basis for the creation of a new triangle element; second, a new mesh node or an existing mesh node is connected to the base segment to generate a new triangle element; last, the generation segment front set and triangle elements are updated. 
	
	Attracted by the classical advancing front method, some researchers have tried to combine it with machine learning modules to design new practical algorithms. Early advancing front method can generate simple meshes, but it involves many inefficient calculations~\cite{lohner1988generation, guo2021improved}. Yao \etal \cite{yao2005ann} combined new vertex position prediction and classification to generate quadrilateral meshes. RLQMG \cite{pan_reinforcement_2022} combined the advancing front method with reinforcement learning, which generated wavefronts and new points by state representation and action representation, respectively. Lu \etal \cite{lu_new_2022} used deep learning to learn to advance direction and step. The main limitation of this type of method is that the quality of the generated mesh cannot be guaranteed (e.g., too many singularities). In addition, the existing advancing front-based methods only generate planar meshes.
	
	\section{Classification based on unit elements}\label{4}

	Based on the type of output mesh primitive, we categorize IMGs into triangular, quadrilateral, hybrid polygon, and tetrahedral mesh generation. This classification allows us to identify the type of mesh generated by each method, thus enabling users to select the content that best suits their needs. As demonstrated in TABLE \ref{tab:classfication}, the majority of current IMG methodologies produce triangular meshes. This trend is logical given that triangular meshes are still the prevailing format, supported by a wealth of tools and theoretical knowledge. This observation also underscores that fields other than triangular meshing, such as quadrilateral and volumetric meshing, are areas ripe for further exploration.
	
	\subsection{Triangular Mesh Generation}\label{4.1}
	Triangle meshes are arguably the most predominant surface mesh representation. Its popularity is derived from its simplicity, flexibility, and the existence of many data structures for efficient mesh navigation and manipulation. With an increase in publicly available triangular mesh datasets and further development of deep learning techniques, many methods have been proposed to generate triangular meshes of given point clouds, images or other forms of data.
	
	Notably, intelligent triangular mesh generation algorithms attracted the interest of researchers as early as the 1990s. The main focus of this period was density-adaptive triangular mesh generation algorithms\cite{chang-hoi_self-organizing_1991,lowther_density_1993,alfonzetti_automatic_1996,alfonzetti_finite_1998,alfonzetti2003neural,alfonzetti_optimized_2008}, whose main feature is the generation of triangular meshes of desired nonuniform density on an initial coarse mesh following the guidance of the density function learned by neural networks. However, the above methods mainly focus on planar meshes and rarely consider the surface mesh generation of 3D objects. Therefore, to improve the practicability of the algorithm to adapt to real-world inputs, more researchers have dedicated themselves to investigating 3D mesh generation. Peng \etal \cite{peng_3d_2004} explored the feasibility of neural networks in triangular mesh reconstruction and 3D object representation. Although this approach is very simple, it realizes lifting from 2D to 3D by predicting the $z$ coordinate.
	
	As 3D data acquisition equipment rapidly develops, triangle mesh generation with 3D data as input is becoming an important branch of IMG. Triangle mesh generation broadly falls into two types: (1) explicit methods, which directly recover a triangle mesh from the input points, and (2) implicit methods, which are aimed at recovering a volumetric function whose zero level-set encodes the surface. For the former, we consider the point cloud as a degenerate mesh (i.e., a mesh without the connection relationship between two points). Therefore, the selection of vertices and prediction of the correct adjacency is needed. Specifically, Scan2mesh \cite{dai_scan2mesh_2019} generated triangle meshes by joint prediction of point positions and connection relations. However, using the fully connected graph as the candidate set of edges makes the edge information redundant. Daroya \etal \cite{daroya_rein_2020} presented REIN, an RNN-based network to generate meshes with varying numbers of vertices by sequential edge prediction. Some methods based on template deformation, as mentioned in Section~\ref{3.1}, are also equivalent to initializing the point connection relations. For the latter, surface triangle meshes are often reconstructed with the marching cubes \cite{lorensen1987marching}, marching tetrahedra \cite{doi1991efficient} or Poisson surface reconstruction \cite{kazhdan2013screened} algorithms. However, these methods cannot maintain sharp features and are not end-to-end. Therefore, some researchers have explored differentiable marching cube algorithms to maintain sharp features. Liao \etal \cite{liao_deep_2018} demonstrated a differentiable alternative to achieve end-to-end training. Liu \etal \cite{liu2021deep} proposed a deep moving least squares method to generate triangular meshes based on iso-surfaces. Chen \etal \cite{chen_neural_2021} redesigned the marching cube algorithm and constructed some tessellation templates, allowing better preservation of sharp features. Chen \etal \cite{chen_neural_2022} improved the dual contouring algorithm\cite{ju2002dual} to eventually allow for end-to-end training.
	
	Another branch is the triangular mesh generation method based on 2D images or features, which is more ambiguous than the method based on 3D data and demands more robustness prior to assist mesh generation. The reconstruction of occlusion information and the estimation of depth information become particularly important. Some deformation-based methods and parameterization-based methods have become the mainstream methods to address the abovementioned cases. For the former, the image is considered the information that guides the deformation of the template. The typical representative is the Pixel2Mesh series\cite{wang2018pixel2mesh,wen2019pixel2mesh++,wang2020pixel2mesh,wang_pixel2mesh_2021}. For the latter, we understand 2D features as a parameterization of 3D object surfaces. The geometry image-based \cite{sinha2017surfnet,li_pcgan_2022} approach deserves to be noted.
	
	\subsection{Quadrilateral Mesh Generation}
	Due to the appealing tensor-product nature and smooth surface approximation of quad meshes, quad meshing techniques are
	usually preferred over triangle meshes in many applications, such as texturing, simulation with finite elements, and B-spline fitting. To take advantage of the powerful representation and generalization capabilities of deep learning, some researchers have also tried to combine deep learning with quad mesh generation and have obtained good solutions \cite{chang-hoi_self-organizing_1991,yao2005ann,smirnov2020learning, pedone_learning_2021,dielen_learning_2021,deng_sketch2pq_2022,chen_mgnet_2022,pan_reinforcement_2022}.
	\begin{figure}[htb]
		\centering
		\includegraphics[width=0.8\linewidth]{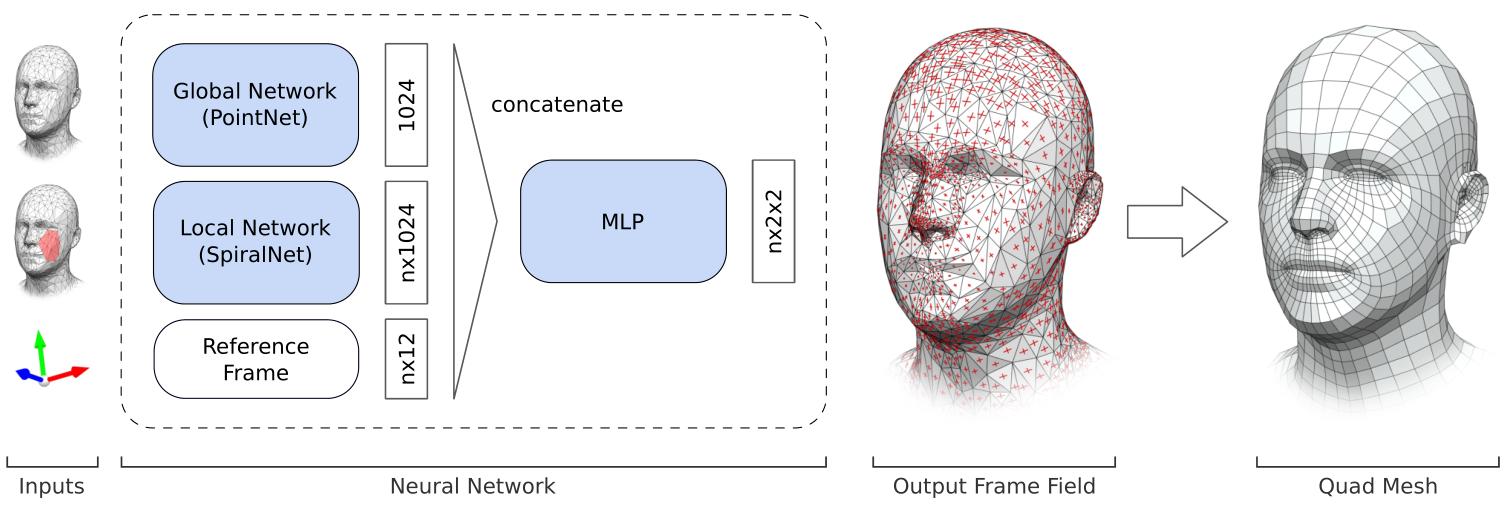}
		\caption{LDFQ\cite{dielen_learning_2021}: an IMG method for quadrilateral mesh.}
		\label{fig:Quadrilateral}
		\vskip -0.1in
	\end{figure}
	
	Following the traditional method of classification, we categorize deep learning-based quadrilateral mesh generation into \textbf{direct} \cite{chang-hoi_self-organizing_1991,yao2005ann, pedone_learning_2021,chen_mgnet_2022,pan_reinforcement_2022} and \textbf{indirect} \cite{smirnov2020learning,dielen_learning_2021,deng_sketch2pq_2022} methods, according to whether intermediate model representation is needed. In these direct IMG methods, Ahn \etal\ \cite{chang-hoi_self-organizing_1991} proposed a self-organizing neural network that achieves automatic nonuniform density quad mesh generation by deforming the initial mesh. Yao \etal\ \cite{yao2005ann} proposed an artificial neural network-based element extraction method for automatic finite element quad mesh generation. The authors designed reasonable node insertion types and then employed neural networks to predict the node positions and insertion types. The mesh boundary is continuously advanced by several iterations of updates. Pedone \etal\ \cite{ pedone_learning_2021} generated a database of small deforming rectangular
	meshes. Chen \etal\ \cite{chen_mgnet_2022} introduced a differential method for structured mesh generation in an unsupervised manner. The method takes boundary curves as input, employs a well-designed neural network to analyze the potential meshing rules, and outputs the mesh with a desired number of cells. To overcome the difficulties of conventional methods in achieving the
	balance between high-quality mesh and computational complexity, Pan \etal\ \cite{pan_reinforcement_2022} proposed a reinforcement learning-based method for automatic quadrilateral mesh generation. Unfortunately, all these methods can only handle simple planes or surfaces.
	
	For indirect methods, Smirnov \etal\ \cite{smirnov2020learning} learned a special shape representation: a deformable parametric template composed of Coons patches. Given a raster image, first, the system infers a set of parametric surfaces that realize the input in 3D. Then, quadrilateral meshes are generated using a template. Deng \etal\ \cite{deng_sketch2pq_2022} presented the Sketch2PQ system, which uses stroke lines, depth samples, and the visible and occluded region masks induced from the sketch as input. First, Sketch2PQ calculates a direction field and B-spline surface as the intermediate model representation. Then, a quad mesh is extracted from the B-spline surface and CDF via geometry optimization. Another interesting work is LDFQ \cite{dielen_learning_2021}, which learns cross fields from a triangular mesh to generate a quadrilateral mesh. Their network can infer frame fields that resemble the alignment of quads. This rich guidance information is able to guarantee the generation of correct and high-quality quad meshes. The pipeline of LDFQ is depicted in Fig. \ref{fig:Quadrilateral}.
	
	\subsection{Hybrid Polygon Mesh Generation}
	In practical applications, due to various nonideal conditions, it is often impossible to generate only a single basic facet. Therefore, some researchers have designed hybrid mesh generation algorithms out of active or passive intention. However, due to the limitation of their application, there are only a few types of IMG with hybrid meshes as the generation target. In the literature that we searched, only BP-ANN \cite{lu_new_2022}, AnalyticMesh \cite{lei_learning_2021} and PolyGen \cite{pmlr-v119-nash20a} belong to this category. BP-ANN generates anisotropic quadrilateral and isotropic triangular meshes by the advancing front method, which improves the level of automation and efficiency of hybrid mesh generation. AnalyticMesh marches among analytic cells to recover the exact mesh of the closed, piecewise planar surface captured by an implicit surface network. The algorithm is applicable to more advanced MLP architectures, including those with shortcut connections and max pooling operations, which support a richer set of architectural designs for learning and exactly meshing complex surface shapes. Polygen more compactly represents the surface of the object with different polygons.

	\subsection{Tetrahedral Mesh Generation}
	Tetrahedral mesh generation is an important branch of mesh generation. Unlike the surface mesh, tetrahedral mesh generation simultaneously divides the surface and the interior of the object. Tetrahedral mesh plays an irreplaceable role in numerical simulation, so it has received the attention of many researchers. However, due to the complexity of the problem and the incompleteness of the basic theory, intelligent tetrahedral mesh generation is still lacking. In the literature that we selected, the tetrahedral mesh generation process has been partially \cite{alfonzetti2003neural} or fully \cite{gao_learning_2020} replaced by deep learning modules. In \cite{alfonzetti2003neural}, Alfonzetti \etal \ proposed a neural network generator for tetrahedral meshes. Starting from an initial moderately coarse mesh, the generator grows the mesh up to the user-specified number of nodes by a node probability density function. DefTet \cite{gao_learning_2020} is optimized for both vertex placement and occupancy and is differentiable with respect to standard 3D reconstruction loss functions.
	\begin{figure}[htb!]
		\centering
		\includegraphics[width=0.8\linewidth]{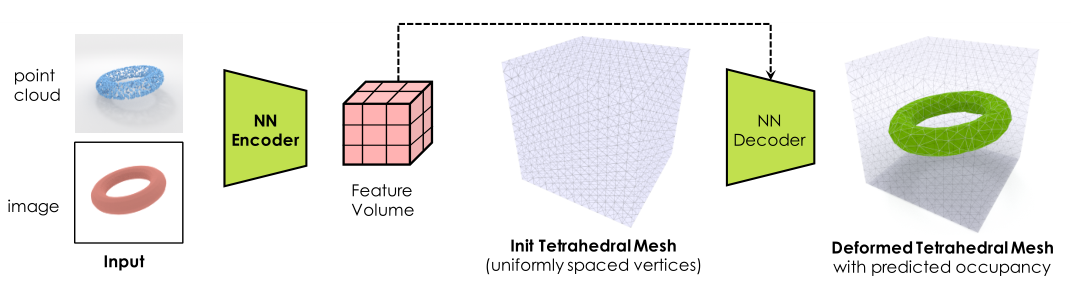}
		\caption{DefTet \cite{gao_learning_2020}: an IMG method for tetrahedral mesh.}
		\label{fig:DefTet}
		\vskip -0.1in
	\end{figure}
	DefTet offers several advantages over prior work and can output shapes with arbitrary topology, using the occupancy of the tetrahedrons to differentiate the interior of the object from the exterior of the object. DefTet also represents local geometric details by deforming the triangular faces in these tetrahedrons to better align with the object surface, thus achieving high-fidelity reconstruction at a significantly lower
	memory footprint than existing volumetric approaches. DefTet accepts point clouds or images as input data to generate meshes. The algorithm flowchart of DefTet is shown in Fig. \ref{fig:DefTet}.

	\section{Classification based on Data types}\label{5}
	According to the type of applicable input data, we categorize IMG into categories such as point cloud-based, image-based, voxel-based, mesh-based, boundary or sketch-based, and latent variable-based mesh generation. This classification allows us to identify the type of input data that each method requires, thereby enabling users to choose the method that is most relevant to their specific data types or use-cases. As illustrated in TABLE \ref{tab:classfication}, the majority of existing IMG methods generate meshes from either images or point clouds, likely due to the relative ease of obtaining these types of data compared to others. Further details are provided in the subsequent subsections.
	
	\subsection{Point Cloud-based Mesh Generation}
	
\begin{figure}[htb]
		\centering
		\includegraphics[width=0.7\linewidth]{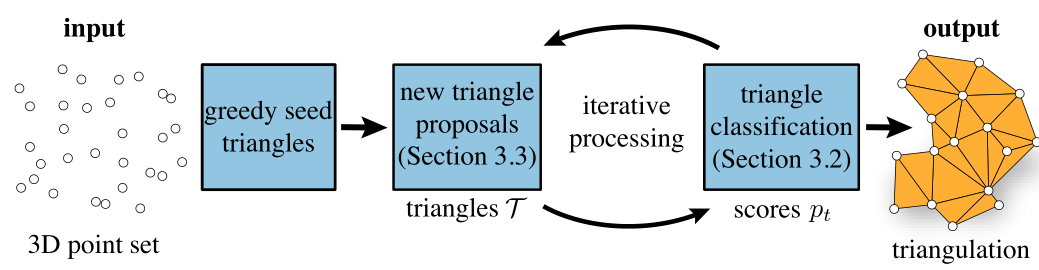}
		\caption{ PointTriNet\cite{sharp_pointtrinet_2020}: a point cloud-based IMG method.}
		\label{fig:point-based}
		\vskip -0.05in
	\end{figure}

	Point clouds, as a major representation of 3D data, have been widely employed in fields such as autonomous driving and AR. Point clouds naturally have depth information and are not affected by ambient lighting conditions compared to RGB images. Reconstructing a mesh from a point cloud is a long-standing problem in computer graphics. Recently, various approaches have been developed to reconstruct shapes for an array of applications \cite{berger2017survey}. These methods are divided into two categories: direct estimation of point locations and connection relationships, as shown in Fig. \ref{fig:point-based}, and implicit surface reconstruction based on isosurfaces. This division is illustrated in section~\ref{4.1}. Next, we introduce these methods in two aspects: (1) the challenges of point cloud-based mesh generation and (2) the learnable priors and related methods proposed to cope with these challenges.
	
	The challenges of point cloud-based IMG methods are derived from both data and mission objectives. For the data side, although point clouds are easily accessible 3D data, they are often sparse and noisy or even incomplete. To address these challenges, researchers have proposed a variety of approaches \cite{li_incomplete_2008,badki2020meshlet,daroya_rein_2020,yifan2021iso,boulch_needrop_2021,ma_neural-pull_2021,ma2022reconstructing,williams2022neural}. Li \etal \cite{li_incomplete_2008} was the first to consider the problem of generating meshes from incomplete point clouds based on neural networks. Badki \etal \cite{badki2020meshlet} handled sparse or noisy point clouds by learning local geometric priors while ensuring the consistency of the global geometry. Boulch \etal \cite{boulch_needrop_2021} and Ma \etal \cite{ma_neural-pull_2021,ma2022reconstructing} reconstructed highly accurate meshes from sparse point clouds by pulling query 3D locations to their closest points on the surface. For the task objective, large scene mesh reconstruction and manifold mesh generation are the focus of this field. Great progress has been made in mesh reconstruction of large scenes \cite{peng2020convolutional,siddiqui2021retrievalfuse,song2021vis2mesh,tang2021sa,sulzer_scalable_2021,williams2022neural}, but it is still a long way from practical application. Manifold mesh generation \cite{rakotosaona2021differentiable,chen_neural_2021,rakotosaona_learning_2021,peng2021shape,chen_neural_2022} is mainly achieved by combining Delaunay triangulation or the Marching Cubes algorithm. For example, Chen \etal \cite{chen_neural_2021} introduced the first Marching Cubes based approach capable of recovering sharp geometric features. Rakotosaona \etal \cite{rakotosaona_learning_2021} leveraged the properties of 2D Delaunay triangulations to construct a 3D mesh. 
	
	\subsection{Image-based Mesh Generation}
	
	The goal of image-based mesh generation is to infer the 3D mesh of objects or scenes from one or multiple 2D images. Recovery of the lost dimensions from just 2D images is an ill-posed problem and fundamental to many applications, such as robot navigation, 3D modeling and animation, industrial control, and medical diagnosis. The rapid development of deep learning techniques, and more importantly, the increasing public datasets, accelerate the evolution of this subfield. Despite being recent, these methods have demonstrated exciting and promising results on various tasks \cite{han2019image}. The following is a detailed introduction.
	\begin{figure}[htb]
		\centering
		\includegraphics[width=0.8\linewidth]{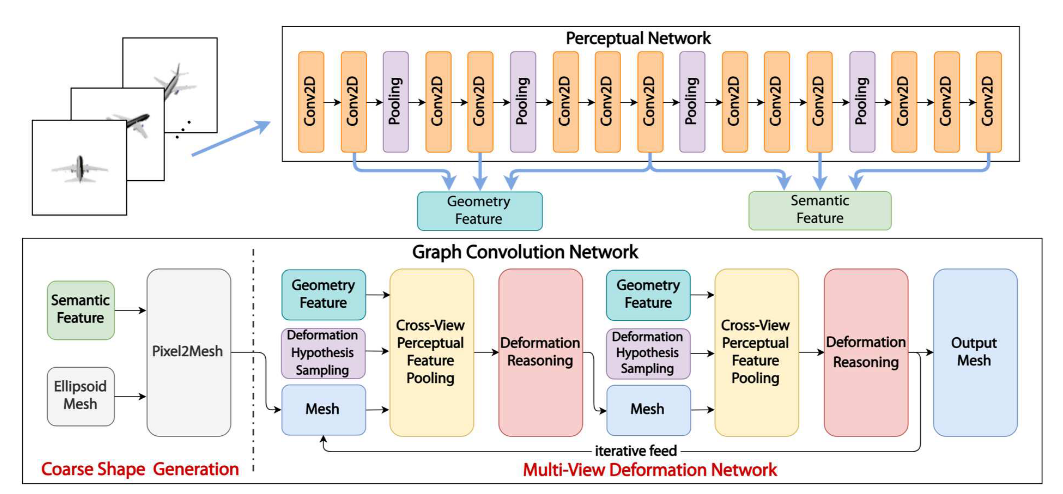}
		\caption{Pixel2Mesh++ \cite{wen2019pixel2mesh++}: an images-based IMG method.}
		\label{fig:image-based}
		\vskip -0.05in
	\end{figure}
	
	One of the immediate difficulties faced by image-based methods is how to maintain and represent the geometric and topological information of objects. Sinha \etal \cite{sinha2017surfnet} constructed networks to learn geometric images with images as input to generate 3D meshes. The geometric information of the shape is preserved by parameterization learning. Wang \etal \cite{wang2018pixel2mesh} used the deformation-based approach to restore geometry and topology. Afterward, there were many deformation-based methods \cite{wen2019pixel2mesh++,tong2020x,li20203d,dongsheng_3d_2020,wang2020pixel2mesh}. Wen \etal \cite{wen2019pixel2mesh++} used multiple images to build geometric details, as shown in \ref{fig:image-based}. Tong \etal \cite{tong2020x} employed specific templates to provide more geometric details. Although there are many excellent works, generalization is still a weakness of these deformation-based IMG technologies.
	
	Other researchers have turned their attention to the reconstruction of textured meshes. Although RGB images provide some texture information, the task is more challenging than purely generating a mesh. Henderson \etal \cite{henderson_leveraging_2020} focused on the problem of generating a 3D texture mesh from 2D images. Munkberg \etal \cite{munkberg2022extracting} simultaneously optimized topology, materials, and illumination from multiview image observations. Admittedly, the interpretability of IMG is enhanced when exploring the correspondence between 3D mesh textures and 2D images. In addition, mesh generation based on image sequences and videos is also worthy of attention\cite{bozic2021transformerfusion,li_learning_2021,pedone_learning_2021,sun2021neuralrecon,yang2021lasr,jiang2022selfrecon}.

	\subsection{Voxel-based Mesh Generation}
	A voxel is a data structure that uses a fixed-size cube as the minimum unit to represent a 3D object. A voxel is a traditional method for storing volume data, providing density, opacity, normals, and other information. The index of the voxel grid provides location information. The major drawback of voxel modeling is that the storage and calculation of voxels require a large memory resource.
	
	Similar to point clouds, a class of voxel-based methods uses occupancy and isosurfaces to generate the mesh implicitly. Mescheder \etal \cite{mescheder2019occupancy} used occupancy networks for 3D resolution enhancement to better reconstruct objects.
	Voxel2Mesh \cite{wickramasinghe2020voxel2mesh} utilized a network to extract features from voxelized MRI brain images and CT liver scans and relied on deformation networks to obtain a triangular mesh of the target object. Peng \etal \cite{peng2020convolutional} combined convolutional
	encoders with implicit occupancy decoders and achieved detailed reconstruction of objects and 3D scenes. The authors encode the input into 2D or 3D Voxel grids which are processed using convolutional networks and then decoded into occupancy probabilities via a fully connected network. DMTet \cite{NEURIPS2021_30a237d1} is a deep 3D conditional generative model that can synthesize high-resolution, 3D shapes using simple user guides such as coarse voxels. The model marries the merits of implicit and explicit 3D representations by a differentiable marching tetrahedra layer. This combination allows joint optimization of the surface geometry and topology as well as the generation of the hierarchy of subdivisions by using reconstruction and adversarial losses.
	\begin{figure}[htb]
		\centering
		\includegraphics[width=0.76\linewidth]{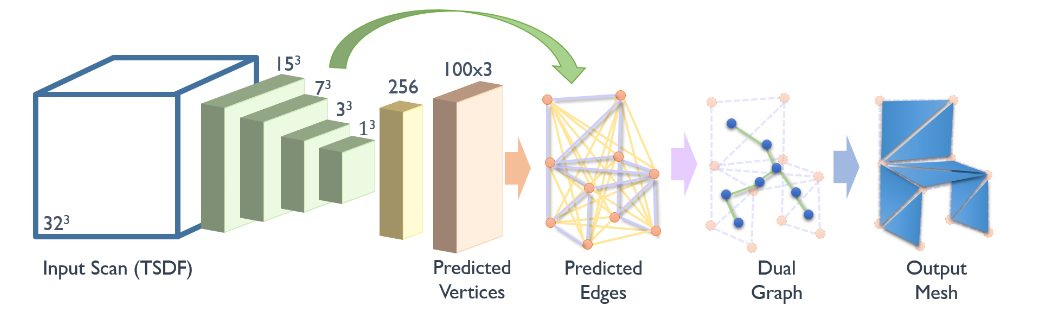}
		\caption{Scan2Mesh\cite{dai_scan2mesh_2019}: a voxel-based IMG method.}
		\label{fig:DF-based}
		\vskip -0.05in
	\end{figure}
	
	As an intermediate product of iso-surface extraction, the voxelized distance field is also often employed as input to generate meshes \cite{dai_scan2mesh_2019,chen_neural_2021,chen_neural_2022}. Scan2Mesh \cite{dai_scan2mesh_2019} took the voxelized truncated signed distance field (TSDF) as input. Its framework is visualized in Fig. \ref{fig:DF-based}. Scan2Mesh is composed of two main components: first, a 3D-convolutional and graph neural network to jointly predict vertex locations and edge connectivity; and second, a graph neural network to predict the final mesh face structure. NMC \cite{chen_neural_2021} used a signed distance field as input and was then trained to reconstruct the zero-isosurface of an implicit field while preserving geometric features such as sharp edges and smooth curves. The main limitations of NMC are that it is sensitive to rotation and cannot avoid self-intersection of meshes. NDC \cite{chen_neural_2022} is a data-driven approach to mesh reconstruction based on dual contouring. NDC can be trained to produce meshes from binary voxel grids, signed or unsigned distance fields, or point clouds and can produce open surfaces in cases where the input represents a sheet or partial surface. 
	
	\subsection{Mesh-based Mesh Generation}
	In IMG, mesh-based input is mainly utilized in two tasks: mesh optimization \cite{alfonzetti_automatic_1996,alfonzetti_finite_1998,alfonzetti2003neural,alfonzetti_optimized_2008,hahner_mesh_2022} and quadrilateral mesh generation \cite{dielen_learning_2021}.
	Early deep learning-based mesh generation tended to take a coarse mesh as input and refine it to meet the geometric and topological information of the object by neural networks. In \cite{alfonzetti_automatic_1996}, Alfonzetti proposed an automatic mesh generator that is the Let-It-Grow neural network. Starting from a rough mesh of triangles with a small number of vertices, this algorithm increases the number of vertices until a user-selected value is reached. The vertex growth is driven by a predefined probability density function. In Alfonzetti's follow-up work \cite{alfonzetti_finite_1998,alfonzetti2003neural,alfonzetti_optimized_2008}, he followed the idea in \cite{alfonzetti_automatic_1996} and proposed different solutions. {   Liu \etal \cite{liu2020neural} proposed a novel framework for data-driven coarse-to-fine geometry modeling. }
	Hahner \etal \cite{hahner_mesh_2022} proposed an autoencoder that handled semiregular meshes of different sizes and topologies. This algorithm can reconstruct meshes with high quality and generalize to the dynamics of unseen time sequences. Dielen \etal \cite{dielen_learning_2021} proposed a neural network that inferred frame fields from unstructured triangle meshes. Afterward, a quad mesh is reconstructed from the frame field by an existing parametrization based quadrangulation method.
	
	\subsection{Boundary- or Sketch-based Mesh Generation}
	Compared to point clouds or images, boundaries and sketches are two types of data that are more concise and easier to obtain. Therefore, boundaries and sketches are popular input data forms for mesh generation tasks. The boundary defines the area covered by the mesh, gives the initial position and growth direction for the mesh and is a common input for advancing front-based mesh generation methods. Sketches depict rich geometric features of 3D shapes, such as silhouettes, occluding and suggestive contours, ridges and valleys and hatching lines, and thus provide a succinct and intuitive method for mesh generation. In the IMG that we collected, boundary-based mesh generation
	methods \cite{chang-hoi_self-organizing_1991,lowther_density_1993,yao2005ann,zhang_meshingnet_2020,lu_new_2022,chen_mgnet_2022,pan_reinforcement_2022} are usually employed for 2D planar meshes, while sketch-based methods \cite{han2017deepsketch2face,lun20173d,li2018robust,yan2020interactive,smirnov2020learning,deng_sketch2pq_2022,du_sanihead_2022} are generally utilized for 3D object surface meshes.
	
	\begin{figure}[htb]
		\centering
		\includegraphics[width=0.85\linewidth]{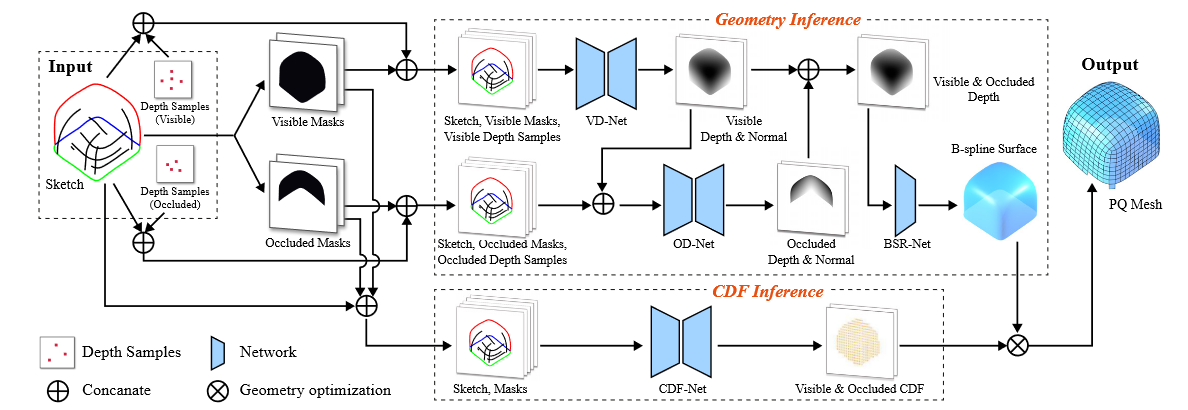}
		\caption{Sketch2PQ\cite{deng_sketch2pq_2022}: a sketch-based IMG for quad mesh.}
		\label{fig:sketch-based}
		\vskip -0.15in
	\end{figure}
	
	For boundary-based IMG, most methods generate 2D meshes by the advancing front technique \cite{yao2005ann,pan_reinforcement_2022,lu_new_2022} or by solving partial differential equations with given initial boundary conditions \cite{zhang_meshingnet_2020, chen_mgnet_2022}. In addition, A. Chang-Hoi \etal \cite{chang-hoi_self-organizing_1991} and D. Lowther \etal \cite{lowther_density_1993} generated meshes by deformation and point prediction, respectively. A sketch can be considered a very sparse image. Therefore, additional information (e.g., depth information) or global and local constraints are needed. Lun \etal \cite{lun20173d} divided sketch-based generation into two stages: first, generating depth and normal images from a multiview sketch and, second, generating a surface from the depth and normal images. Li \etal \cite{li2018robust} used CNNs to infer the depth and normal maps that represent the surface, with an intermediate layer that models the curvature direction field and produces a confidence map to improve robustness. Deng \etal \cite{deng_sketch2pq_2022} generated quadrilateral meshes using visible and occluded boundaries, contour, and feature lines, and some depth samples. The boundary and the feature lines collectively determine the style of mesh, as shown in \ref{fig:sketch-based}.
	
	\subsection{Latent Variable-based Mesh Generation}
	As the two dominant generation models, GANs and variational autoencoders (VAEs) \cite{kingma2013auto} naturally exert their power in IMG. It is well known that both models can generate images directly from latent variables (random noise) after training. Inspired by the success in image generation, some researchers have conducted mesh generation research based on GANs and VAEs. Considering that only a latent variable of latent space is needed to generate mesh after model training, this paper classifies this kind of method as latent variable-based mesh generation \cite{ranjan2018generating,bouritsas2019neural,li_pgan_2019,wu2020pq,yuan_mesh_2020,li_pcgan_2022}. Specifically, Li \etal \cite{li_pgan_2019,li_pcgan_2022} used the GAN model to generate geometry and normal images from a latent variable and then generated a mesh by postprocessing. Some other works \cite{ranjan2018generating,bouritsas2019neural,yuan_mesh_2020} selected the VAE model to reconstruct, deform, or interpolate the mesh. Pq-Net \cite{wu2020pq} enabled the sequence generation of parts using a conditional variational autoencoder.
	\begin{figure}[htb]
		\centering
		\includegraphics[width=0.85\linewidth]{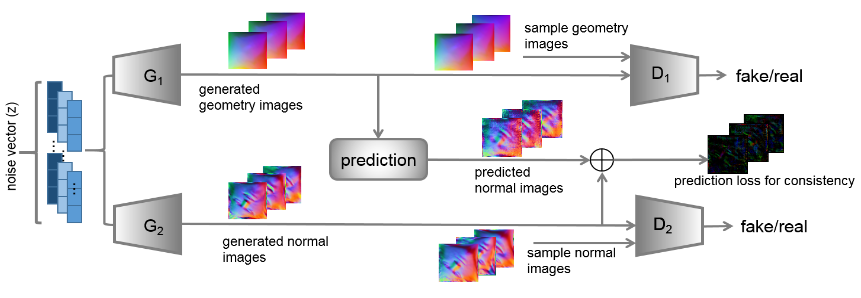}
		\caption{PGAN\cite{li_pgan_2019}: a latent variable-based IMG method.}
		\label{fig:noise-based}
		\vskip -0.15in
	\end{figure}
	
	\section{Evaluation}\label{6}
	
	Generally, evaluation methodologies for various IMGs fall into two major categories: qualitative and quantitative. Our qualitative analysis focuses on the content perspective of all IMG methods, including targeting challenges, advantages, limitations, and average monthly citations. On the other hand, quantitative evaluations typically emphasize specific metrics such as time complexity and mesh quality. In this section, we undertake a thorough and comprehensive qualitative analysis based on the extracted content from all selected methods. However, due to the vastly different target tasks and datasets across various IMG methods, the metrics they employ differ significantly. Consequently, we refrain from a quantitative comparison of all IMG methods, instead providing a list of commonly used metrics.
	
	\subsection{Content evaluation}
	We extracted the key challenges, advantages, and limitations of the proposed algorithm from the reviewed papers and sorted them in Table~\ref{tab:procons} in chronological order. The key challenges are often the goal of a paper and embody its value to a certain extent. Advantages are often reflected in effective solutions to challenges. The limitations point out the shortcomings of the algorithm and indicate directions for future research. In specific applications, these three aspects also provide guidance for us to choose suitable algorithms.
	
	In the summary of advantages and limitations, we did not use phrases such as higher accuracy, faster speed, or fewer memory requirements. With the development of IMG technology, the advantages claimed in the original paper are often untenable. Therefore, we try to give a more objective evaluation based on the mission objectives and application scenarios. In addition, we calculated and displayed the average monthly citations (AMC) of each paper, which to some extent reflects the impact of the corresponding work.
	\subsection{Metrics}\label{Metrics}

	In our early attempt at quantitative analysis, we experienced complicated and arduous data collation and analysis. Although we ultimately failed, we also obtained an unexpected discovery that the lack of benchmarks is a challenge hindering the development of IMG. The benchmark here includes unified test data and metrics. For quantitative analysis, most of the methods create their own test data. Even if the dataset is the same, the test data will vary due to different preprocessing. Moreover, the difference in metric selection also directly leads to the inability of quantitative comparisons. Due to the above insurmountable issues, we abandoned the attempt to make a comprehensive quantitative comparison of existing IMG methods. Next, we list and explain the commonly employed metrics in IMG so that readers can understand the challenges of comprehensive comparison in this field. Based on what metrics evaluate, we classified these metrics into error and feature preservation metrics, topological metrics, and perceptual metrics.
	
	\noindent\textbf{Error and feature preservation metrics}. 
	In IMG, the chamfer distance (CD) \cite{fan2017point} is the most prevalent metric to measure the error between the reconstructed mesh $\mathcal{M}_{gt}$ and the target surface $\mathcal{M}_{pred}$. The CD is defined as the average of the minimum distance between two point clouds. The coordinate range of points, density of points, and definition of the distance norm will directly affect the value of the CD. Whether or not the coordinates are normalized, the number of sampling points on the mesh, and whether the distance is the $L_1$ or $L_2$ norm, yield different CD values, making comparative analysis impossible.
	F score \cite{wang2018pixel2mesh} is defined as the harmonic mean of precision and recall, where recall is a fraction of points on $\mathcal{M}_{gt}$ that lie within a certain distance to $\mathcal{M}_{ped}$, and precision is the fraction of points on $\mathcal{M}_{ped}$ that lie within a certain distance to $\mathcal{M}_{gt}$. The F score mainly captures significant mistakes and disregards small mistakes that significantly contribute to visual artifacts. However, different threshold and distance selection strategies directly lead to different F score values.
	The normal consistency score \cite{mescheder2019occupancy} measures the consistency of the normal of the 3D model surface, that is, the normals of adjacent faces should be smooth and continuous, rather than obviously discontinuous or inconsistent. The test data must contain accurate normal vectors and involve complex projection, so the application scenario is limited.
	
	\noindent\textbf{Topological metrics}.
	Watertightness and manifoldness are two topological metrics. Water tightness \cite{sharp_pointtrinet_2020} denotes the percentage of edges that have exactly two incident triangles, and manifoldness \cite{sharp_pointtrinet_2020} represents the percentage of edges with one or two incident triangles. These two metrics measure the mesh from the perspective of topology but disregard the geometry and features of the target surface.
	
	\noindent\textbf{Perception metric}.
	As \cite{sorkine2003high,chen_learning_2019} remarks, metrics such as the mean squared error and CD do not account for the visual quality of the object surfaces. For visual perception, low-frequency errors in shapes are less noticeable than high-frequency errors. To overcome the shortcomings of traditional quality metrics, deep learning measurement methods based on visual similarity or perceptual quality \cite{abouelaziz_curvature_2016, abouelaziz_convolutional_2017, abouelaziz_reduced_2018,abouelaziz_convolutional_2018, abouelaziz2021learning,chen2020developing,chen2021mve} have emerged. However, the generalization and applicability of these models need to be improved, so they are rarely employed in quantitative comparative analysis.
	
	As mentioned above, we cannot use the existing results to compare all these IMG methods, nor can we compare them directly by experiments. It is very challenging to conduct a comparative analysis of these IMG methods. Nevertheless, to facilitate researchers in choosing papers of their own interest, we have provided some imperfect qualitative comparison results and comparative relationship diagrams in the supplementary materials and the project repository.
	
	\section{Peripheral}\label{7}
	\subsection{Common Datasets}
	Data serves as the cornerstone of IMG development. The variety of inputs underscores the diversity of the datasets employed for IMG methodologies. Regrettably, public datasets for volume meshes are currently non-existent. Thus, our focus here is limited to commonly utilized datasets for triangle and quadrilateral meshes.

	\noindent\textbf{Triangle mesh}. These commonly employed triangle mesh datasets include Princeton ModelNet \footnote[1]{http://modelnet.cs.princeton.edu/}, ShapeNet \cite{chang2015shapenet}, TOSCA \cite{bronstein2006efficient}, COSEG \cite{wang2012active}, surface reconstruction benchmark of Berger \cite{berger2013benchmark} and Williams \cite{williams_deep_2019}, Thingi10K \cite{zhou2016thingi10k}, D-FAUST \cite{bogo2017dynamic}, Famous \cite{erler_points2surf_2020}, and the CAD dataset ABC \cite{koch2019abc}; a dataset for single image 3D shape modeling: Pix3d \cite{sun2018pix3d}; facial expression datasets: COMA \cite{ranjan2018generating} and MeIn3D \cite{booth20163d}; human body shapes datasets: MGN \cite{bhatnagar2019multi} and MultiHuman \cite{zheng2021deepmulticap}; clothed body meshes with real texture: RenderPeople \cite{renderpeople}, Axyz \cite{axyz2019}, and Digit Wardrobe \cite{bhatnagar2019multi}; and indoor scenes datasets: ScanNet \cite{dai2017scannet}, Scenenet \cite{handa2016scenenet}, Matterport3d \cite{chang2017matterport3d},
	
	{
		\setlength\LTleft{0pt}
		\setlength\LTright{0pt}
		\centering
		\onecolumn
		\tiny
		\linespread{1.22} \selectfont
		\begin{longtable}{p{2cm}|p{4cm}p{5cm}p{4.5cm}c}
			\caption{Challenges, advantages and limitations of all selected IMG methods. AMC denotes average monthly citations.}\\
			\bottomrule 
			Article & Challenges & Advantages  &  Limitations & AMC \\
			\hline
			\endhead
			\rowcolor{Ocean} Self-organizing \cite{chang-hoi_self-organizing_1991}& Nonuniform density mesh generation& Generates meshes with a specified density distribution & Only simple cases with given boundary are handled & 0.12\\
			Lowther \etal \cite{lowther_density_1993}&Nonuniform density mesh generation & Totally automatic,density adaptive mesh  & Results extremely irregular, boundary needed & 0.08\\
			\rowcolor{Ocean} Alfonzetti \etal \cite{alfonzetti_automatic_1996}& Nonuniform density mesh generation& Density adaptive mesh; boundary preserving & Require an initial rough mesh & 0.12 \\
			Alfonzetti \etal \cite{alfonzetti_finite_1998}& Nonuniform density mesh generation & The vertex density function can be adaptively calculated & Require an initial rough mesh & 0.07\\
			\rowcolor{Ocean} Alfonzetti \etal \cite{alfonzetti2003neural} & Nonuniform tetrahedral mesh generation & Can generate a preset number of tetrahedral meshes & A rough initial mesh and density function are needed & 0.04\\
			Peng \etal \cite{peng_3d_2004}&Hand and represent the complex 3D object & Combine multilayer perception with 3D Object Reconstruction& The MLP only needs to recover the $Z$ coordinate & 0.14\\
			\rowcolor{Ocean} Yao \etal \cite{yao2005ann}&Formulate mesh elements extraction rules for IMG &Reduces the number of singularities to a certain extent &Not end-to-end; may fall into a local minimum&0.16\\
			Alfonzetti \etal  \cite{alfonzetti_optimized_2008}& Generate high quality meshes for 3D cases. & Good quality of the output meshes; simplicity of the algorithm &Only simple cases with given boundary are handled &0.04\\
			\rowcolor{Ocean} Li \etal \cite{li_incomplete_2008}&Robust to incomplete input data& Incomplete points cloud data surface reconstruction & Neural networks are only used to complement point clouds & 0.02\\
			Agostinho  \etal \cite{de_medeiros_brito_junior_adaptive_2008}& Detailed multiresolution mesh generation& In a multiresolution fashion; only a simple initial mesh are needed & Only applicable to genus 0 object mesh generation &0.25\\
			\rowcolor{Ocean} Wen \etal \cite{wen_rbf_2009}& Ill-conditioned matrix and overfitting problem in RBF method & Overcome the numerical ill-conditioning of coefficient matrix and overfitting problem & The selection of the radial basis function limits the solution space of the implicit function&0.05\\
			Xiong \etal \cite{xiong_robust_2014}& Overcome the limitations of multistage processing& Jointly optimize geometry and connectivity; robust to outliers and noise; preserve sharp features &Convergence is not guaranteed; local minimum cannot be avoided; may fail in the hole regions &0.69\\
			\rowcolor{Ocean} DeepGarment \cite{danerek_deepgarment_2017}&Intelligent, efficient and practical IMG& Efficient capture of 3D garment shapes from a single image only. & Only simple T-shirts and dresses can be handled. &0.89\\
			Deepsketch2face \cite{han2017deepsketch2face}&Convert sketches to meshes efficiently & infer meshes from 2D sketches; effectively help users create face meshes. & Only for caricature model; Cannot create details such as wrinkles; &1.74\\
			\rowcolor{Ocean} ShapeMVD \cite{lun20173d}&Efficiently convert sketches to meshes & Output meshes preserve topology and shape structure& Output meshes lack details; not end-to-end differentiable&2.47 \\
			Surfnet \cite{sinha2017surfnet}& Directly generates meshes of rigid and nonrigid shapes& Solves the problem of area distortion to a certain extent. & Only genus 0 surfaces can be generated &2.60 \\
			\rowcolor{Ocean} Pixel2mesh \cite{wang2018pixel2mesh}&Output meshes lose surface details &Certainly maintains surface details  & Only object similar to the initial mesh topology can be reconstructed&16.82\\
			AtlasNet \cite{groueix2018papier}& Generate high-resolution, 3D shapes & Generate shapes of arbitrary resolution; broad application & Exist distortion or overlap &16.29 \\
			\rowcolor{Ocean} Li \etal \cite{li2018robust}& Modeling generic freeform 3D surfaces from sparse, expressive 2D sketches. & A freeform surface modeling; just needs succinct line annotations; solves the flow field by CNN & Cannot create geometric details; Model one patch each time; not end-to-end differentiable &1.22\\
			DMC \cite{liao_deep_2018}&End-to-end marching cubes algorithm&End-to-end watertight meshes generation & High memory requirements and restricted to $32^3$ voxel resolution &3.04\\
			\rowcolor{Ocean} CoMA \cite{ranjan2018generating}&Generate 3D meshes of human faces & Generate diverse realistic 3D faces & Require a reference template from the same object class & 6.60\\
			3D-CFCN  \cite{cao2018learning}& Incomplete and noise input with occlusions& To a certain extent, robust to noise &Poor mesh generation for thin slices &0.50\\
			\rowcolor{Ocean} MGN \cite{bhatnagar2019multi}& Predict object geometry & Mesh reconstruction with texture captured from images& Cannot handle pose dependent deformations; rely on segmentation&5.03\\
			3DN \cite{wang20193dn}&Simple and efficient IMG algorithm &An end-to-end optimizable model & Require a reference template with the same topology &2.23 \\
			\rowcolor{Ocean} TMN  \cite{pan2019deep}&High quality mesh generation &A wide variety of topological meshes can be reconstructed & Unable to directly generate open surfaces &3.12 \\
			ONet \cite{mescheder2019occupancy}& Simple and efficient mesh representation& Can extract 3D meshes at any resolution; can handle various inputs &High memory requirements for 3D voxel grid & 28.79\\
			\rowcolor{Ocean} N3DMM \cite{bouritsas2019neural}& Efficient feature representation for mesh & Create anisotropic mesh; lightweight and easy-to-optimize model& Require a reference template from the same object class & 2.48\\
			PGAN \cite{li_pgan_2019}& Explore geometric images for mesh generation & Generate diverse, realistic, 3D faces& Complex pre/post-process; mesh distortion by parameterization&0.03\\
			\rowcolor{Ocean} HumanMeshNet \cite{venkat_humanmeshnet_2019}& Real-time reconstructions &Ensuring smooth surface reconstruction& Training data and test data must be the same basic model &0.35\\
			DISN \cite{xu_disn_2019}&High-quality and detailed mesh generation& Capture fine-grained details and generate high-quality 3D models & Only able to handle objects with a clear background &8.45\\
			\rowcolor{Ocean} IM-Net \cite{chen_learning_2019}& Generate high visual quality mesh & Efficient representation with coordinate information & Many nonsurface points yield many invalid calculations &17.10\\
			Scan2Mesh \cite{dai_scan2mesh_2019}&Adaptation to incomplete scans& Cleaner and more CAD-like meshes from noisy and partial range scans & High memory requirements; does not enforce mesh regularity or continuity &1.67\\
			\rowcolor{Ocean} DGP \cite{williams_deep_2019}& Dense reconstruction of the input point cloud & Robust to noise; capture sharp features; arbitrary resolution meshes. & High computational complexity; no adaptive patch selection &2.59\\
			Mesh R-CNN \cite{gkioxari_mesh_2019}&High-quality and detailed mesh generation& Applicable to unconstrained real-world images with many objects& Nonmanifold, low reconstruction accuracy & 8.03\\
			\rowcolor{Ocean} DeepSDF \cite{park2019deepsdf}& Trade-offs across fidelity, efficiency and compression capabilities &Enables high-quality shape representation, interpolation and completion from partial and noisy 3D input data& Simple fully connected network architecture without local information or translation equivariance &32.23\\
			Pixel2mesh++  \cite{wen2019pixel2mesh++}& Fusion of information across views &Exploit cross-view information; Strong generalization &Only genus 0 surface can be generated &3.74\\
			\rowcolor{Ocean} PQ-Net \cite{wu2020pq}& Encoding of local structure and local geometry &Learn a 3D shape representation in the form of sequential part assembly& Cannot learn symmetry relations or produce topology-altering meshed&2.15\\
			BCNet \cite{jiang2020bcnet}&Accuracy of geometric representation &Different topology garments can be generated &Some details cannot be generated &2.48\\
			\rowcolor{Ocean}   PolyGen \cite{pmlr-v119-nash20a} &Generate a wide variety of realistic geometries&Is capable of generating coherent and diverse mesh samples&It is suitable for the generation of artificial object surfaces but not for the generation of complex surfaces.&3.68\\
			DGTS \cite{hertz2020deep} &Learn the geometric information of the grid &Transferring geometric information between two surfaces with different topologies; no parameterization needed &Inability to effectively learn the overall semantic information; high requirements for input data& 1.30\\
			\rowcolor{Ocean}   Neural Subdivision \cite{liu2020neural}&  Generate smooth and feature-preserving subdivision results.&  Complex nonlinear subdivision schemes can be learned &  Global semantic information cannot be guaranteed &1.83\\
			Mobile3drecon \cite{yang2020mobile3drecon} & Real-time mesh generation &Real-time dense mesh reconstruction& Poorly maintained sharp features  &3.33\\
			\rowcolor{Ocean} Sal \cite{atzmon2020sal}&Applicable to unsigned geometric data&Not needed for ground truth normal data or inside/outside labeling&Bad generation for thin parts &7.30\\
			Pixel2mesh2 \cite{wang2020pixel2mesh}&Preserve surface details &Geometric regularization constrains&Only for genus 0 surface &0.76\\
			\rowcolor{Ocean} Voxel2mesh \cite{wickramasinghe2020voxel2mesh}&No postprocessing needed&Can generate 3D meshes without any postprocessing &Just suitable for genus 0 surface generation &1.39\\
			X-ray2shape \cite{tong2020x}&Applicable to low-contrast image data &Effective for low-contrast image as input & Too many singularities; cannot handle complex topology &0.30\\
			\rowcolor{Ocean} Li \etal \cite{li20203d}& Background reconstruction&The final mesh less affected by the environment & Preprocessing determines mesh quality; thin part is poorly generated &0.02\\
			Meshlet \cite{badki2020meshlet}&Robust to sparse and noisy, 3D points&Valid for sparse and noisy data; regular and globally consistent meshes&Failure on thin structures; the resolution is fixed &1.08 \\
			\rowcolor{Ocean} LIG \cite{jiang2020local}&IMG for scale and complexity indoor scenes&Can generate meshes for arbitrary scale scenes with details&Assume that different categories share similar part geometries&7.85 \\
			CONet \cite{peng2020convolutional}&Scalable and complex scenes mesh generation& Integrate local and global information; obtain translation equivariance&Not applicable to sparse point cloud and not robust to noise &13.59 \\
			\rowcolor{Ocean} ILSM \cite{yan2020interactive}&Realistic liquid splashes mesh generation & Generate realistic liquid splash meshes from simple user sketch input&Unable to generate detailed splatters as real-world splatters; weak generalization& 0.22\\
			IER \cite{liu_meshing_2020}&Generate fine-grained details; good generalizability&Preserve details; handle ambiguous structures; good generalizability&Nonmanifold mesh with holes; need posttreatment for holes&0.83 \\
			\rowcolor{Ocean} Yang \etal \cite{dongsheng_3d_2020}&Generate object details & End-to-end architecture& Only applicable to genus 0 shape & 0\\
			MeshingNet \cite{zhang_meshingnet_2020} &Automatic unstructured mesh generation. & A nonuniform mesh generation method; Combine PDEs with IMG & Only planar mesh generation was tested & 0.81\\
			\rowcolor{Ocean} Point2Surf \cite{erler_points2surf_2020} &Handle partial scan and noise. & Applicable to nonuniform noisy input, objects with varying topological & Not an end-to-end differentiable process &2.73\\
			DEFTET \cite{gao_learning_2020} &Single image to tetrahedral meshes generation & Applicable to any topology; Efficient algorithm; no postprocessing &With flipped tetrahedrons; feature not guaranteed&1.38\\
			\rowcolor{Ocean}BTM \cite{rios_back_2020}&Generalizability of single mesh prototype &More generalizable than comparable methods&High computational complexity, multiple mesh prototypes needed&0.10\\
			PointTriNet \cite{sharp_pointtrinet_2020}&Efficient, scalable and generalizable algorithms & Trained in an unsupervised manner, robustness to outliers & Mesh with many holes, nonmanifold &1.04\\
			\rowcolor{Ocean} Surface Hof \cite{wang_surface_2020}&High-resolution surface reconstruction  &Generate detailed mesh at arbitrary resolution for various topologies& Only for images with a clear background; not end-to-end differentiable&0.10\\
			REIN \cite{daroya_rein_2020} &Mesh generation with sparse point cloud& Mesh generation for sparse point clouds,  & Nonmanifold, no permutation invariance &0.04\\
			\rowcolor{Ocean} Henderson \etal \cite{henderson_leveraging_2020} & Produce textured meshes &No self-intersections; does not need ground-truth segmentation masks& Require a reference template with the same topology &1.83\\
			SSRNet\cite{mi_ssrnet_2020} &Large-scale point clouds mesh reconstruction&Strong scalability; good at reconstructing geometry details & Not an end-to-end differentiable process; the partition method is crucial &1.26\\
			\rowcolor{Ocean} MeshVAE \cite{yuan_mesh_2020} & An effective pooling operation of mesh & Pooling operation based on mesh simplification; can generate details& Only for uniform mesh; fail with nonwatertight or irregular mesh&0.65\\
			Point2Mesh \cite{hanocka_point2mesh_2020} & Learnable priori for mesh generation & Robust to unoriented normals and noise; produce watertight meshes; unsupervised learning &Initial mesh needed; high computational complexity and low reconstruction accuracy &3.68\\
			\rowcolor{Ocean} NMF \cite{gupta_neural_2020}&Manifold mesh generation & Generate two-manifold meshes & Only for genus 0 shapes &1.26\\
			NDF \cite{chibane2020neural}&High-resolution outputs of arbitrary shape&Can generate open surface;can represent inner structures &Need postprocessing for meshing  &3.67\\
			\rowcolor{Ocean} Meshsdf \cite{remelli2020meshsdf}&A differentiable way to produce surface mesh& End-to-end differentiable; watertight mesh for arbitrary topology& Insufficient generalization performance, only generate simple meshes&2.19\\
			Smirnov \etal \cite{smirnov2020learning}&Efficient IMG.&Final mesh with fewer singularities &Postprocessing needed &0.87\\
			\rowcolor{Ocean} Lasr \cite{yang2021lasr}&Nonrigid structures mesh generation&Does not need category-specific mesh template; good generalization &Fails at heavy occlusions; efficiency needs improvement &1.47\\
			Transformerfusion \cite{bozic2021transformerfusion}& Efficient coding and accurate reconstruction &Online reconstruction approach running at interactive frame-rates &Cannot construct details for some scenes &2.36\\
			\rowcolor{Ocean} CSPNet \cite{venkatesh2021deep}& Efficient and low memory representation for complex surfaces &Can represent complex surfaces of any topology; efficient computation of local geometric properties &Not an end-to-end differentiable process&0.82\\
			DASM \cite{wickramasinghe2021deep}& Utilization of local regularization &A plug-and-play smoothing module can generate smoother meshes.& Hand-designed metric needed for mesh smooth&0.47\\
			\rowcolor{Ocean} Neuralrecon \cite{sun2021neuralrecon}& Accuracy, Consistency, and Real-time reconstruction &Can generator accurate and coherent reconstruction in real-time&Not an end-to-end differentiable process&2.93\\
			Sa-convonet \cite{tang2021sa}& Scalability to large-scale scenes; Generalizability& Valid for large scenes& Slow inference speed &1.18\\
			\rowcolor{Ocean} DMTet \cite{NEURIPS2021_30a237d1}&High-quality and detailed mesh from IMG&Can generate meshes with arbitrary topology, finer geometric details, fewer artifacts& Global uniform resolution; suffers from bad local minimum; tends to produce double/broken surfaces &0.81 \\
			Vis2mesh \cite{song2021vis2mesh}& Large-scale scenes mesh generation &Good generalization; robust to noise; detail reconstruction &The system is complex &0.09\\
			\rowcolor{Ocean} IMLSNet  \cite{liu2021deep}& Transform the discrete point sets into smooth mesh &Define implicit functions on point sets&Not an end-to-end differentiable process&1.60\\
			Retrievalfuse \cite{siddiqui2021retrievalfuse}& Mesh reconstruction of large scenes& Accurate scene reconstructions; maintain local detail&The results of the retrieval may be suboptimal &1.27\\
			\rowcolor{Ocean} Deepdt  \cite{luo2021deepdt}& Inside/outside classification cannot obtain clean meshes& Without ground truth labels of tetrahedrons or visibility information&  Unable to handle excessively large input&0.47\\
			Iso-points \cite{yifan2021iso}&IMG for noisy and incomplete input&Faster convergence and accurate recovery of details and topology&Not explicitly model the appearance&0.87\\
			\rowcolor{Ocean} DST \cite{rakotosaona2021differentiable}& The differentiable structure of triangular mesh & Control both the vertex positions and the topology; linear complexity to the number of vertices& Surface partition needed; visible artifacts exist across boundaries; cannot handle numerous points or patches &0.38\\
			SAP \cite{peng2021shape}&Efficient and differentiable point to mesh layer&Output watertight manifold meshes; interpretable, lightweight and short inference time; robustness to outliers & Limited to small scenes; cubic memory requirements &1.60\\
			\rowcolor{Ocean} Bertiche \etal \cite{bertiche_deep_2021}&Generalizability to various geometry and topology& Allows continuous predictions with differential geometry & Missing garments details in the final mesh &0\\
			NeeDrop \cite{boulch_needrop_2021}&Mesh generation with extremely sparse point cloud& A self-supervised method;input data can be sparse point clouds& Requires postprocessing to generate meshes  &0.33\\
			\rowcolor{Ocean} AnalyticMesh \cite{lei_learning_2021}& {  Accurate} mesh reconstruction& Doe not lose the details of the implicit field compared to the Marching Cube &postprocessing, such as smoothing or holes filling are needed&0\\
			LMR \cite{li_learning_2021}&Mesh inference with in-the-wild videos& Local-dynamical-modeling approach to video mesh recovery& Generated mesh cannot match the human body &0\\
			\rowcolor{Ocean} NRSfM \cite{pedone_learning_2021}&Nonrigid surface reconstruction & A simple and efficient model for nonrigid deformed mesh generation& The generated mesh may overlap at patch boundaries&0.11\\
			DHSP \cite{wei_deep_2021}&Mesh generation with sparse colored point cloud; multipriori integration& Generate mesh with high-resolution texture; robust to sparse and noisy point clouds & Complex structure and operation; only single object models with simple topology can be processed &0.27\\
			\rowcolor{Ocean} Neural-Pull \cite{ma_neural-pull_2021}&Learning high-quality SDFs to generate mesh &Learn SDF without the signed distance value; robust to noise & Missing sharp features in the reconstruction mesh &1.30\\
			DI-Fusion \cite{huang_di-fusion_2021} &The tradeoff between memory and mesh quality & Simultaneously encode geometric and uncertain information & Unable to preserve spatial continuity sharp features&1.47\\
			\rowcolor{Ocean} DSE \cite{rakotosaona_learning_2021}&Generate manifold meshes & Produces near-manifold triangulations, robustness to outliers & Not E2E differentiable; require alignment for patches &0.67\\
			LDFQ \cite{dielen_learning_2021}& Most of the IMG methods of quadrilateral rely on a dense user-provided direction field & A robust data-driven approach to the computation of direction fields&The way the ground truth of the frame field is constructed determines the quality of the mesh &0.09\\
			\rowcolor{Ocean} DGNN \cite{sulzer_scalable_2021}&IMG with Large-scale scenes and incomplete point cloud input& Mesh reconstruction for large-scale, defect-laden point clouds; take into account visibility
			information & Low generalization, limit precision when the acquisition is noisy, missing reconstruction details &0.17\\
			Hu \etal \cite{hu_mesh_2021}&Generation of complex geometric features& Achieve a trade-off between mesh density and feature representation & The final mesh has more overlap &0.07\\
			\rowcolor{Ocean} NMC \cite{chen_neural_2021} &IMG method that persevere sharp features&Can maintain sharp geometric features and learning local topology& Sensitive to rotation; possible self-intersection in the generated mesh&0.89\\
			Skeletonnet \cite{tang2021skeletonnet}&Lack of constraints on the topology. & Topology-preserved; high-quality skeletal volume& Cannot deal with natural images in the wild; high algorithm complexity&0.80\\
			\rowcolor{Ocean} Nvdiffrec \cite{munkberg2022extracting} &Joint optimization of topology, material and lighting & An appearance-aware and end-to-end mesh generator with materials. & High calculation and memory consumption; simplified shading model&3.33 \\
			Selfrecon \cite{jiang2022selfrecon}&Combine the advantages of implicit and explicit representations& Produce high-fidelity surfaces for arbitrarily clothed humans with self-supervised optimization& Long optimization time; mainly suitable for self-rotating motions&2.00\\
			\rowcolor{Ocean} Autosdf \cite{mittal2022autosdf}&Powerful priors for mesh generation & Useful for
			multimodal generation for a diverse set of tasks& Sensitive to alignment; CAD models only; not end-to-end differentiable&3.67\\
			NKF \cite{williams2022neural}&Generate large scenes mesh with sparse point cloud; generalizability&Can reconstruct shape categories outside the training set; can reconstruct large scenes mesh& Kernel implementation requires a dense linear solve; requirement of oriented points &2 \\
			\rowcolor{Ocean} Sketch2PQ \cite{deng_sketch2pq_2022}&Real-time IMG& Real-time mesh generation with a dense direction field & Not E2E differentiable; only for disk topology surfaces&0\\
			SRMAE \cite{hahner_mesh_2022}&Good generalizability & Can handle meshes in different sizes; suitable for unbounded mesh &Cumbersome preprocessing; underutilization of global information &0.25\\
			\rowcolor{Ocean} OnSurfacePrior \cite{ma2022reconstructing}&IMG with sparse point cloud and valid prior & SDF and normals are not needed; applicable to sparse point clouds& Not an end-to-end differentiable process &1.33\\
			Lu \etal \cite{lu_new_2022}&Intelligent advancing-front method; & An automation advancing front based IMG method& Valid only for 2D mesh; no control over singularities&0\\
			\rowcolor{Ocean} MGNet\cite{chen_mgnet_2022}&Differential structured mesh generation & Unsupervised structured quad mesh generation; requires no prior knowledge or measured dataset& Boundary accuracy determines the quality of the mesh; only for planar mesh &0.33\\
			RLQMG \cite{pan_reinforcement_2022}&Efficient and intelligent quad-mesh generation. & Automatic quad-mesh generation without extra clean-up operations & Lack of control over singularities; only for planar mesh generation &0.20\\
			\rowcolor{Ocean} TopoNet \cite{ben_charrada_toponet_2022}&Topological generalizability & Not limited by the topology of the template; better geometry capture& The reconstructed meshes with small holes&0\\
			SAniHead \cite{du_sanihead_2022}&Mesh generation of animal heads & Can generate meshes with geometric details& Hard to create shapes with thin structures and poor generalization ability &1.33\\
			\rowcolor{Ocean} NDC \cite{chen_neural_2022}&Difficult to obtain surface gradients for Dual Contouring method& Good reconstruction quality; applicable to various inputs & Produce nonmanifold mesh, not completely invariant to orientation &0\\
			PCGAN \cite{li_pcgan_2022}&Process topology similarity
			among meshes by CNN&Preserve topology information and the spatial structure& Need preprocess and postprocess for mesh generation &0\\
			\rowcolor{Ocean} Nice-slam \cite{zhu2022nice}&Large scenes oversmooth mesh generation&Can fill small holes and extrapolate scene geometry; real-time system &Coarse representation only & 4.33\\
			NRGBD\cite{azinovic2022neural}&Scenes high-quality mesh reconstruction of room-scale scenes &Effective combination of depth observations and neural radiance field for room-scale scene reconstruction& Need postprocessing; lack of details; slow convergence & 8.67\\
			\rowcolor{Ocean} POCO \cite{boulch2022poco}&Scalability of implicit reconstruction& Suitable for both single-object reconstruction and whole-scene reconstruction& Cannot complete shapes when missing large parts; without normals, yield orientation failures&2.00
			\label{tab:procons}\\
			\hline
		\end{longtable}%
		
	}
	\begin{twocolumn}
		
		\noindent  and Synthetic Rooms \cite{peng2020convolutional}. The two most frequently used datasets are ShapeNet and ModelNet. ShapeNet contains 3D models from a multitude of semantic categories and organizes them under the Word-Net taxonomy. ShapeNet has indexed more than 3,000,000 3D models, 220,000 of which are classified into 3,135 categories. ModelNet has 662 object categories and 127915 CAD models. It contains three subsets: Modelnet10 with 10 categories; Modelnet40 with 40 categories; and Aligned40 with 40 classes aligned 3D model.
		
		\noindent\textbf{Quadrilateral mesh}. The commonly used quadrilateral mesh datasets include QuadWild \cite{pietroni2021reliable}; Clothed body meshes with real texture: RenderPeople \cite{renderpeople}, Axyz \cite{axyz2019}. Notably, the quadrilateral mesh datasets are not sufficient, which also hinders the development of this field to a large extent.
		
		\subsection{Commonly Used Prior}
		In IMG, commonly employed underlying priors include \emph{smoothness}, \emph{primitives}, \emph{distribution}, \emph{user-driven}, \emph{orientation}, \emph{visibility} and \emph{regularity} priors. In the absence of supervision information, these priors provide effective guidance for model optimization. Below, we introduce them in detail.
		
		The surface \emph{smoothness} prior constrains the reconstructed surface to satisfy a certain level of smoothness. The most general form is local smoothness, e.g., the Laplace smooth constraint, which often serves as a regular term of the objective function  \cite{wang2018pixel2mesh,gkioxari_mesh_2019,wen2019pixel2mesh++,jiang2020bcnet,gao_learning_2020,dongsheng_3d_2020,wickramasinghe2020voxel2mesh,wang2020pixel2mesh,wickramasinghe2021deep}. Geometric regularization \cite{ma2022reconstructing,deng_sketch2pq_2022} attempts to pull points to their neighbor. With smooth prior constraints, the model often filters out noise but also loses some high-frequency details.
		
		The geometric \emph{primitive} prior assumes that the scene geometry may be explained by a compact set of simple geometric shapes, i.e., planes, boxes, spheres, and cylinders.

		{  The \emph{orientation} prior is an important prior in traditional mesh generation and is applicable to deep learning-based mesh generation tasks. The normal consistency prior guarantees the consistency of orientation when generating object surfaces based on distance fields \cite{atzmon2020sal}.} 
		
		The \emph{visibility} prior makes assumptions about the exterior space of the reconstructed scene and how this can provide cues for combating noise, nonuniform sampling, and missing data. Specifically, Deng \etal\ used \emph{visibility} in Sketch2PQ \cite{deng_sketch2pq_2022} by a depth-normal compatibility term and depth sample term. Sulzer \etal \cite{sulzer_scalable_2021} skillfully utilized visibility information derived from camera positions and produced watertight meshes. Song \etal \cite{song2021vis2mesh} explicitly employs depth completion for the visibility prediction of 3D points.
		
		The global \emph{regularity} prior takes advantage of the notion that many shapes possess a certain level of regularity in their higher-level composition. Regularity refers to the relationship between the whole and the part \cite{wu2020pq,jiang2020local}. The same kind of object is composed of similar basic subparts in a similar arrangement. For example, cars are composed of wheels, chassis, and cover from bottom to top. In addition, symmetry is also a common regularity.
		
		{  \subsection{Classical mesh-based learning architectures}}
		With the success of deep learning methods in computer vision, many neural network models have been introduced to conduct 3D shape representation using volumetric grids \cite{maturana2015voxnet,brock2016generative,wang2017cnn,park2019deepsdf,chen2021decor} and point clouds \cite{qi2017pointnet,qi2017pointnet++,yang2018foldingnet,thomas2019kpconv,Wang2019Dynamic,li2022weakly,wiersma2022deltaconv}. However, due to the complexity and irregularity of mesh data, there are only a few neural networks that simultaneously use the geometric and topological information as inputs \cite{hanocka2019meshcnn,feng2019meshnet,milano2020primal,singh2021meshnet++}. MeshCNN \cite{hanocka2019meshcnn}, a convolutional neural network designed for triangular meshes, defined edge-centric convolution and pooling operations. Convolutions are applied on edges and the four edges of their incident triangles, and pooling is applied via an edge collapse operation that retains the surface topology. These operations facilitate a direct analysis of mesh in its native form. MeshNet \cite{feng2019meshnet} defined face-centric convolution and pooling operations for triangular meshes. To synthesize geometric information and topological information, the network provides an effective data preprocessing and network framework. PD-MeshNet \cite{milano2020primal} utilized features for both edges and faces of a 3D mesh as input and dynamically aggregated them using an attention mechanism. Singh \cite{singh2021meshnet++} designed surface correlation blocks that capture local features at various scales.
		
		Although only a few of the network models mentioned above can simultaneously process geometry and topology, many outstanding works have emerged in the research of mesh analysis and feature representation. Convolutions that exploit mesh connectivity structures were introduced by Masci \etal \cite{masci2015geodesic}. Since then, several approaches have addressed the irregularity of the mesh structure and sampling rate, proposing ideas such as uniformly sampling the neighborhood of each vertex or achieving regularity by spectral decomposition \cite{boscaini2016learning, lim2018simple,schult2020dualconvmesh,zhou2020fully,smirnov2021hodgenet}. Recently, combined with a learned heat diffusion operation, DiffusionNet \cite{sharp2022diffusionnet} offered a unified perspective across representations of surface
		geometry. Another line of work seeks to understand the structure and connectivity of the mesh \cite{gupta_neural_2020,yuan_mesh_2020,zhou2020fully}. The authors implemented convolutional layers over point features related to vertices, for example, in EdgeConv\cite{Wang2019Dynamic}. By combining neighborhood or edge information, these methods can achieve a better comprehensive representation of the geometric and topological information of the mesh surface.
		
		\vskip 0.1in
		\section{Discussion and Conclusion}\label{8}
		IMG significantly enhances the generalizability, robustness, and practical utility of traditional mesh generation techniques. However, it comes with its own set of challenges: dealing with non-ideal data, algorithm practicality, and complex object generation. The non-ideal data challenge entails working with 2D images lacking 3D information, sparse and non-uniform point clouds, occluded or incomplete point clouds, noisy data, and large-scale input data. Algorithm practicality presents hurdles such as ensuring broad generalization, achieving high computational efficiency, maintaining low memory requirements, exhibiting robustness against non-ideal inputs, and guaranteeing end-to-end differentiability. The challenge of generating meshes for complex objects pertains to intricate objects or scenes. This includes the creation of meshes for large and dynamic scenes, high density detail, texture based meshes, water splash meshes, non-uniform density meshes, and structured meshes.
		
		\subsection{Summary of open challenges}
		In this review, we have highlighted the limitations found across the selected articles. These limitations not only represent the challenges within the IMG field, but also help identify potential future research directions. The primary challenges we've recognized are as follows:\\
		(1) \textbf{Generalization:} Generalization means the ability for an algorithm to handle objects not observed during training - is a crucial aspect demonstrating the practical utility of the algorithm. The lack of generalization is a prevalent issue with current IMG methods, particularly those using deformation-based mesh generation techniques. These methods often produce meshes specific to a particular object type or topology, limiting their wider applicability.\\
		(2) \textbf{Interpretability:} The interpretability of IMG is critically important, especially within the aerospace industry. It is our suggestion that melding IMG with fundamental mathematical and physical models, as evidenced by studies such as Zheng et al.~\cite{zheng2021quadrilateral} and Lei et al.~\cite{lei2019geometric}, can not only effectively govern model optimization but also impart enhanced interpretability to the model. However, thus far, no research has directly addressed this particular issue.\\
		(3) \textbf{Lack of Benchmark:} 
		As discussed in Section~\ref{Metrics}, performing a comprehensive comparative analysis of all these IMG methods poses significant challenges. The lack of standardized testing procedures, coupled with variations in test data and metrics, has significantly impeded IMG's advancement. Therefore, the establishment of benchmarking standards would substantially expedite the progress of IMG.\\
		(4) \textbf{Watertight and manifold:} 
		Watertight and manifold meshes are frequently required in many computer graphics applications. However, current element classification based IMG methods cannot generate watertight or manifold meshes. In attempting to generate a mesh from scratch, these methods overlook certain local constraints. Furthermore, due to the discrete nature of watertight and manifold meshes, they fail to provide gradients essential for model optimization.\\
		(5) \textbf{Structured mesh:} 
		Structured meshes offer numerous advantages that unstructured meshes cannot match, including highly efficient storage and access. Unfortunately, due to their inherent complexity, MGNet \cite{chen_mgnet_2022} is currently the only IMG method that considers structured mesh generation. Thus, it is critical to explore more practical structured or semi-structured IMG algorithms capable of handling real-world objects. Additionally, there is a noticeable scarcity of quadrilateral and volume mesh datasets, which hinders the advancement of IMG in areas of structured quadrilateral mesh and volume mesh generation. Creating open-source datasets in these fields could significantly expedite the development of IMG in the context of structured meshes.\\
		(6) \textbf{Textured mesh:} 
		Textured mesh is a highly sought-after representation for 3D objects and finds extensive application in numerous fields, including industrial design and digital entertainment. Existing IMG research has primarily concentrated on the challenges of reconstructing geometry and topology. In contrast, only a handful of studies have focused on the generation of textured meshes \cite{henderson_leveraging_2020, munkberg2022extracting, gao2021tm}. Furthermore, the generation of textured meshes undeniably enhances the interpretability of the model, making this an important area for future exploration in the IMG field.\\
		(7) \textbf{Large or dynamic scene mesh generation:} 
		Scene mesh generation holds immense potential in diverse fields, such as autonomous driving, indoor robotics, and mixed reality. For IMG methods to truly prove useful in real-world applications, several critical properties must be addressed. Firstly, real-time functionality of the algorithm is desired. Secondly, the algorithm should have the capacity to make plausible predictions for regions devoid of observations. Furthermore, the system should demonstrate scalability to accommodate large scenes. Lastly, the robustness against noisy or incomplete observations is essential. Existing methodologies \cite{venkat_humanmeshnet_2019,yang2020mobile3drecon,sun2021neuralrecon,tang2021sa,song2021vis2mesh,li_learning_2021,siddiqui2021retrievalfuse,zhu2022nice,williams2022neural}, however, fall short of satisfying these criteria simultaneously, resulting in scene meshes that tend to be overly smooth and lacking in detail.

		\subsection{Conclusion}
		In this article, we have provided a systematic and comprehensive review of Intelligent Mesh Generation (IMG) methodologies, highlighting core techniques, application scopes, learning goals, data types, targeted challenges, strengths, and limitations. We scrutinized 113 research papers for a detailed analysis and data extraction. Our primary points of extraction included 1) the addressed challenges, 2) the fundamental concepts, application range, advantages, and disadvantages, 3) the type of input data, the nature of output mesh, and mesh quality, and 4) potential directions for future research. Articles were classified based on their techniques, unit elements, and applicable data types. While there have been substantial advancements in IMG in recent years, we also identified numerous issues and challenges that pave the way for future exploration and investigation in this field.
		
		To the best of our knowledge, this is the most recent and comprehensive survey of existing IMG methods. This survey provides a holistic perspective and extensive research resources for scholars in the field of IMG. However, it is worth noting that our review has certain limitations. Specifically, our focus was solely on IMG; non-machine-learning mesh generation methods were beyond our purview. As a result, numerous valuable papers on traditional mesh generation were not included. Furthermore, despite the application of systematic literature review methodology and a manual search to ensure comprehensive inclusion, there may have been relevant papers that were overlooked in the initial selection due to the sheer volume of existing literature.
		
		\section*{Acknowledgment}
		This research was supported by the National Key R$\&$D Program of China (2021YFA1003003) and the National Natural Science Foundation of China under Grants 61936002 and  T2225012.

		\bibliographystyle{IEEEtran}
		\bibliography{bibliography2}
		
		\begin{IEEEbiography}[{\includegraphics[width=1in,height=1.25in,clip,keepaspectratio]{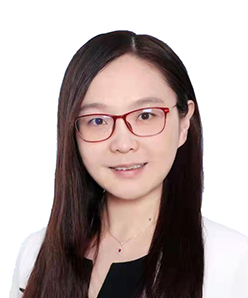}}]{Na Lei}
			received her B.S. degree in 1998 and a Ph.D. degree in 2002 from Jilin University. Currently, she is a professor at Dalian University of Technology. Her research interest is the application of modern differential geometry and algebraic geometry to solve problems in engineering and medical fields. She mainly focuses on computational conformal geometry, computer mathematics, and its applications in computer vision and geometric modeling.
		\end{IEEEbiography}
		
		\begin{IEEEbiography}[{\includegraphics[width=0.90in,height=1.20in,clip,keepaspectratio]{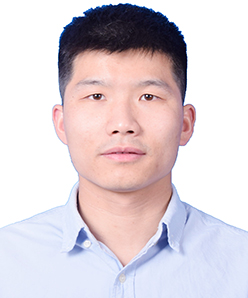}}]{Zezeng Li}
			received a B.S. degree from Beijing University of Technology (BJUT) in 2015. He is currently pursuing a Ph.D. degree from Dalian University of Technology (DUT). His research interests include image processing, point cloud processing, and mesh generation.
		\end{IEEEbiography}
		
		\begin{IEEEbiography}[{\includegraphics[width=0.90in,height=1.20in,clip,keepaspectratio]{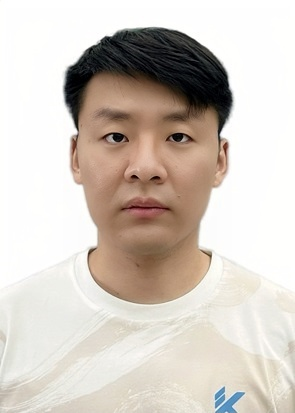}}]{Zebin Xu}
			received a B.S. degree from 
			Dalian University of Technology (DUT)
			in 2020. He is currently pursuing a Ph.D. degree from Dalian University of Technology (DUT). His research interests include image processing and mesh generation.
		\end{IEEEbiography}
		
		\begin{IEEEbiography}[{\includegraphics[width=0.90in,height=1.20in,clip,keepaspectratio]{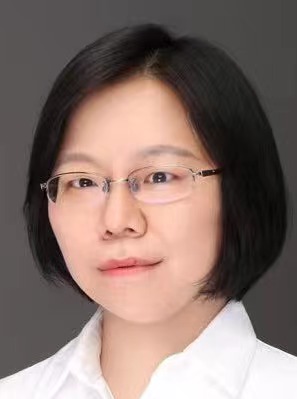}}]{Ying Li}
			received a B.S. degree in 1999 and a Ph.D. degree in 2004 from Jilin University. She completed her postdoctoral research at Tsinghua University in 2007. Currently, she is an associate professor at Jilin University. Her research interests are the application of machine learning models in computational biology and image processing.
		\end{IEEEbiography}

		\begin{IEEEbiography}[{\includegraphics[width=1in,height=1.25in,clip,keepaspectratio]{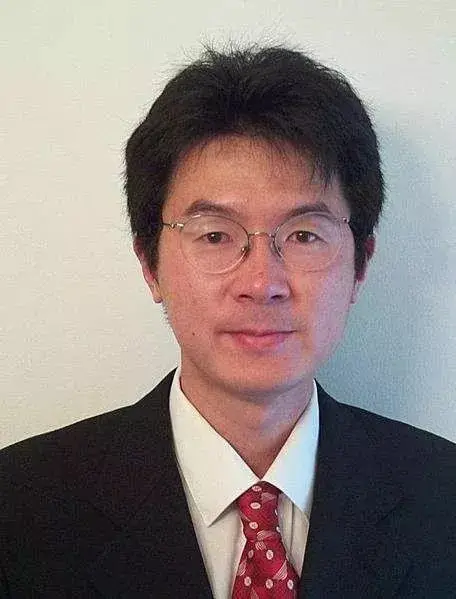}}]{Xianfeng Gu}
			received a B.S. degree from Tsinghua University and a Ph.D. degree from Harvard University. He is now a tenured professor in the Department of Computer Science and Applied Mathematics at the State University of New York at Stony Brook. He has won several awards, such as the NSF CAREER Award of the USA, the Chinese Overseas Outstanding Youth Award, the Chinese Fields Medal, and the Chenxing Golden Prize in Applied Mathematics. Professor Gu's team combines differential geometry, algebraic topology, Riemann surface theory, partial differential equations, and computer science to create a cross-disciplinary "computational conformal geometry", which is widely utilized in computer graphics, computer vision, 3D geometric modeling and visualization, wireless sensor networks, medical images, and other fields.
		\end{IEEEbiography}
		
	\end{twocolumn}
\end{document}